\documentclass[journal]{IEEEtran}

\title{A Scopus-Based Scoping Review of Semantic Web Technologies for Transport Decarbonisation}
\date{\today}

\author{Joseph~D.~Preece~\orcidlink{0000-0002-1854-3578}, Navya~Nayak~\orcidlink{0009-0009-0227-8544}, Haoruo~Zhao~\orcidlink{0000-0003-1960-6515}, Daniel~Mitchell~\orcidlink{0000-0002-9390-4150}, Wenshuo~Tang~\orcidlink{0000-0003-0996-6306}, Duncan~Guthrie~\orcidlink{0009-0002-0105-0408}, Heather~Steele~\orcidlink{0000-0003-0970-0663}, David~Flynn~\orcidlink{0000-0002-1024-3618},~\IEEEmembership{Senior~Member,~IEEE}, Philip~Greening~\orcidlink{0000-0001-7153-4788}, Bernadin~Namoano~\orcidlink{0000-0002-6159-5250}, and Paul~Harvey~\orcidlink{0000-0003-1243-938X}%
\thanks{J. D. Preece, H. Zhao, D. Mitchell, W. Tang, D. Guthrie, D. Flynn, and P. Harvey are with the University of Glasgow, UK.}%
\thanks{N. Nayak and B. Namoano are with Cranfield University, UK.}%
\thanks{H. Steele is with Durham University, UK.}%
\thanks{P. Greening is with Heriot-Watt University, UK.}%
\thanks{Corresponding author: J. D. Preece (email: \href{mailto:joseph.preece@glasgow.ac.uk}{joseph.preece@glasgow.ac.uk})}}

\usepackage{draftwatermark}
    \SetWatermarkText{PREPRINT}
    \SetWatermarkScale{1}

\usepackage{amssymb} 
    
\usepackage[%
    backend=biber,
    style=ieee,
    sorting=none
]{biblatex} 
\usepackage{booktabs} 

\usepackage[showdow, en-GB]{datetime2} 

\usepackage[hidelinks]{hyperref} 
    \usepackage{doi}
    \usepackage{url}
    
\usepackage[prefix]{glossaries-extra} 
    \glsdisablehyper
    \newabbreviation{3lhmm}{3-LHMM}{Three-layer Holistic Maritimization Model}
    \newabbreviation{abox}{ABox}{assertional box}
    \newabbreviation{ai}{AI}{artificial intelligence}
    \newabbreviation{api}{API}{application programming interface}
    \newabbreviation{bim}{BIM}{building information modelling}
    \newabbreviation{cbr}{CBR}{case-based reasoning}
    \newabbreviation{cld}{CLD}{Challenge Led Demonstrator}
    \newabbreviation{credit}{CRediT}{Contributor Roles Taxonomy}
    \newabbreviation{csv}{CSV}{comma separated value}
    \newabbreviation{cv}{CV}{computer vision}
    \newabbreviation{dl}{DL}{description logic}
    \newabbreviation{em}{EM}{electro-mobility}
    \newabbreviation[prefix={the\space}]{epsrc}{EPSRC}{Engineering and Physical Sciences Research Council}
    \newabbreviation[prefix={the\space}]{eu}{EU}{European Union}
    \newabbreviation{ev}{EV}{electric vehicle}
    \newabbreviation{evca}{EVCA}{Urban Electric Vehicle Charging Activity}
    \newabbreviation{fair}{FAIR}{findable, accessible, interoperable, and reusable}
    \newabbreviation{gdp}{GDP}{gross domestic product}
    \newabbreviation{gis}{GIS}{geographic information system}
    \newabbreviation{guid}{GUID}{globally unique identifier}
    \newabbreviation{genclon}{GenCLOn}{Generic City Logistics Ontology}
    \newabbreviation{gof4r}{GoF4R}{Governance of the Interoperability Framework for Rail and Intermodal Mobility}
    \newabbreviation{gps}{GPS}{Global Positioning System}
    \newabbreviation{html}{HTML}{Hypertext Markup Language}
    \newabbreviation{http}{HTTP}{Hypertext Transit Protocol}
    \newabbreviation{iea}{IEA}{International Energy Agency}
    \newabbreviation{ieee}{IEEE}{Institute of Electrical and Electronics Engineers}
    \newabbreviation{ies4}{IES4}{Information Exchange Standard 4}
    \newabbreviation{ifopt}{IFOPT}{Identification of Fixed Objects in Public Transport}
    \newabbreviation{iot}{IoT}{internet of things}
    \newabbreviation{ipoves}{I-POVES}{Interoperable-Pathfinder, Order, Vehicle, Environment, and Supervisor}
    \newabbreviation{imrad}{IMRaD}{Introduction, Methods, Results, and Discussion}
    \newabbreviation{iri}{IRI}{Internationalized Resource Identifier}
    \newabbreviation{iso}{ISO}{International Organization for Standards}
    \newabbreviation{it}{IT}{information technology}
    \newabbreviation{kg}{KG}{knowledge graph}
    \newabbreviation{kgc}{KGC}{\glsxtrlong{kg} construction}
    \newabbreviation{kpi}{KPI}{key performance indicator}
    \newabbreviation{lda}{LDA}{Latent Dirichlet Allocation}
    \newabbreviation{ldf}{LDF}{Linked Data Fragments}
    \newabbreviation{llm}{LLM}{large language model}
    \newabbreviation{lod}{LOD}{linked open data}
    \newabbreviation{lode}{LODE}{Live \glsxtrshort{owl} Documentation Environment}
    \newabbreviation{lrg}{LRG}{local and regional government}
    \newabbreviation{mdpi}{MDPI}{Multidisciplinary Digital Publishing Institute}
    \newabbreviation{nlp}{NLP}{natural language processing}
    \newabbreviation[longplural={points of interest}]{poi}{PoI}{point of interest}
    \newabbreviation{prisma}{PRISMA}{PiattafoRme cloud Interoperabili per SMArt-government}
    \newabbreviation{nhs}{NHS}{National Health Service}
    \newabbreviation{oops}{OOPS!}{OntOlogy Pitfall Scanner!}
    \newabbreviation{om}{OM}{Ontology of Units of Measure}
    \newabbreviation{osm}{OSM}{OpenStreetMap}
    \newabbreviation[prefix={the\space}]{owl}{OWL}{Web Ontology Language}
    \newabbreviation{owl-el}{OWL\,2\,EL}{\glsxtrshort{owl} Existential Language Profile}
    \newabbreviation{prov-o}{PROV-O}{Provenance Ontology}
    \newabbreviation{rag}{RAG}{retreival augmented generation}
    \newabbreviation{rdf}{RDF}{Resource Description Framework}
    \newabbreviation{rdfs}{RDFS}{\glsentryshort{rdf} Schema}
    \newabbreviation{r2rml}{R2RML}{RDB to \glsxtrshort{rdf} Mapping Language}
    \newabbreviation{rml}{RML}{\glsxtrshort{rdf} Mapping Language}
    \newabbreviation{rest}{REST}{Representational State Transfer}
    \newabbreviation{rss}{RSS}{\glsxtrshort{rdf} Site Summary or Really Simple Syndication}
    \newabbreviation{sdg}{SDG}{Sustainable Development Goal}
    \newabbreviation{shacl}{SHACL}{Shapes Constraint Language}
    \newabbreviation{skos}{SKOS}{Simple Knowledge Organization System}
    \newabbreviation{snomed-ct}{SNOMED CT}{Systematized Nomenclature of Medicine -- Clinical Terms}
    \newabbreviation{sparql}{SPARQL}{SPARQL Protocol and RDF Query Language}
    \newabbreviation{srac}{SRAC}{scenario-risk-accident chain ontology}
    \newabbreviation{stub}{STUB}{System for Ticketing Using Blockchain}
    \newabbreviation{sumo}{SUMO}{Simulation of Urban MObility}
    \newabbreviation{swrl}{SWRL}{Semantic Web Rule Language}
    \newabbreviation{tbox}{TBox}{terminological box}
    \newabbreviation{tfidf}{TF-IDF}{Term Frequency-Inverse Document Frequency}
    \newabbreviation{tko}{TKO}{Transdisciplinary Knowledge Ontology}
    \newabbreviation{tm}{TM}{traffic management}
    \newabbreviation{uk}{UK}{United Kingdom}
    \newabbreviation{uml}{UML}{Unified Modelling Language}
    \newabbreviation[prefix={the\space}]{un}{UN}{United Nations}
    \newabbreviation{uob}{UoB}{University of Birmingham}
    \newabbreviation{uri}{URI}{Uniform Resource Identifier}
    \newabbreviation{usa}{USA}{United States of America}
    \newabbreviation{vlr}{VLR}{voluntary local review}
    \newabbreviation{vru}{VRU}{vulnerable road user}
    \newabbreviation{voc}{VOC}{volatile organic compound}
    \newabbreviation{vos}{VOS}{Visualisation of Similarities}
    \newabbreviation{w3c}{W3C}{World Wide Web Consortium}
    \newabbreviation{widoco}{WIDOCO}{Wizard for DOCumenting Ontologies}
    \newabbreviation{wp}{WP}{Work Package}
    \newabbreviation{www}{WWW}{World Wide Web}
    \newabbreviation{xml}{XML}{eXtensible Markup Language}
    \makeglossaries

\usepackage{lipsum} 

\usepackage{orcidlink} 

\usepackage{soul} 

\usepackage{listings} 
    \lstdefinelanguage{list}{%
        alsoletter={-},
        morekeywords=[1]{TITLE-ABS-KEY, DOCTYPE, SUBJAREA, PUBYEAR, LANGUAGE},
        morekeywords=[2]{OR, AND, NOT, W/5, W/10, *},
        keywordstyle=[1]\color{blue},
        keywordstyle=[2]\color{orange},
        sensitive=false,
        morestring=[b]",
        stringstyle=\color{teal},
    }
    \lstdefinestyle{list}{%
        language=list,
        basicstyle=\ttfamily\footnotesize,
        columns=fullflexible,
        keepspaces=true,
        showstringspaces=false,
        breaklines=true,
    }
    \lstdefinelanguage{scopus}{%
        alsoletter={-},
        morekeywords=[1]{TITLE-ABS-KEY, DOCTYPE, SUBJAREA, PUBYEAR, LANGUAGE},
        morekeywords=[2]{OR, AND, NOT, W/5, W/10, *},
        keywordstyle=[1]\color{blue},
        keywordstyle=[2]\color{orange},
        sensitive=false,
        morestring=[b]",
        stringstyle=\color{red},
    }
    \lstdefinestyle{scopus}{%
        language=scopus,
        basicstyle=\ttfamily\footnotesize,
        columns=fullflexible,
        keepspaces=true,
        showstringspaces=false,
        breaklines=true,
    }

\usepackage[version=4]{mhchem} 

\usepackage{multirow}

\usepackage[detect-all]{siunitx} 
    \DeclareSIUnit{\dollar}{\$}
    \DeclareSIUnit{\million}{million}
    \DeclareSIUnit{\trillion}{trillion}

\usepackage{tabularx} 

\usepackage{tikz} 

\usepackage[colorinlistoftodos, disable]{todonotes}

\usepackage[normalem]{ulem} 

\usepackage[table]{xcolor}

\usepackage{cleveref} 
    \crefname{enumi}{RQ}{RQs}
    \Crefname{enumi}{Research Question}{Research Questions}

\newcommand{\region}[1]{Region~\textbf{#1}}
\newcommand{\regions}[1]{Regions~\textbf{#1}}

\newcommand{\scopusterm}[1]{\texttt{#1}}

\begin{document}

\maketitle

\begin{abstract}
Transport is a core component of modern economies, producing approximately \SI{6.9}[\dollar]{\trillion} of global \glsxtrlong{gdp}. However, it also contributes a fifth of total global greenhouse gas emissions. Decarbonising transport is therefore crucial and urgent for meeting climate change goals, yet progress towards these targets is frustrated by fragmented data, heterogeneous systems, and limited interoperability across operational, infrastructural, environmental, and policy domains. Semantic technologies can address these challenges by providing structured, machine-interpretable representations supporting data integration and reasoning. In particular, Semantic Web technologies provide a standardised approach through \glsfmtlong{w3c} specifications. This paper provides a scoping review of the literature at the intersection of semantic technologies, transport, and decarbonisation, with a focus on the adoption of Semantic Web technologies. Using a filtering approach based on Scopus indexed keywords, we refine an initial corpus of \num{222} publications to \num{60} papers for detailed analysis. We find that decarbonisation remains minimally addressed, with only seven papers positioned at the full intersection of semantic technologies, transport, and decarbonisation, and rarely as the leading motivation. We also find that \SI{67}{\percent} of the in-scope studies adopt \glsfmtlong{w3c} standards; however, this adoption does not clearly translate into reproducibility or reuse, as only \SI{22}{\percent} of such studies provide accompanying artefacts. Finally, we note that digital twins are emerging as a notable application trend. Overall, this paper identifies substantial opportunities for future research, particularly as advances in Semantic Web technologies create new possibilities for interoperable, reusable, and reasoning-enabled transport decarbonisation systems.
\end{abstract}


\section{Introduction} \label{sec:introduction}

Transport is a core component of a modern economy, enabling the movement of people, goods, and services. In 2021, the transport sector produced \SI{7}{\percent} of global \gls{gdp}~\cite{partnershiponsustainablelowcarbontransportTransportClimateSustainability2025} (approximately \SI{6.9}[\dollar]{\trillion}~\cite{worldbankGDP}) and employed \SI{193}{\million} people worldwide~\cite{partnershiponsustainablelowcarbontransportTransportClimateSustainability2025}. Investment in transport equipment in \pgls{eu} accounted for \SI{1.9}{\percent} of \gls{gdp} in 2023~\cite{eurostatKeyFiguresEuropean2025}, while households in \pgls{eu} allocated on average \SI{12}{\percent} of their consumption expenditure to transport~\cite{eurostatKeyFiguresEuropean2025}.

Despite these economic and social benefits, transport also plays a significant role in the rapid change of our planet's climate, by releasing greenhouse gases into the atmosphere. In 2023, \SI{15.9}{\percent} of global greenhouse gases and \SI{21.9}{\percent} of global \ce{CO_2} emissions came from transport: predominantly road vehicles~\cite{partnershiponsustainablelowcarbontransportTransportClimateSustainability2025, internationaltransportforumITFTransportOutlook2023}. In 2025, \SI{95.4}{\percent} of energy used in transport came from fossil fuels~\cite{partnershiponsustainablelowcarbontransportTransportClimateSustainability2025}, with fossil fuel subsidies (both implicit and explicit) totalling around \SI{7.2}[\dollar]{\trillion}~\cite{partnershiponsustainablelowcarbontransportTransportClimateSustainability2025}.

As such, the decarbonisation of transport -- the process of reducing or eliminating greenhouse gas emissions released into the atmosphere by the sector -- is crucial and urgent. \Pgls{un} seek to address this via \glspl{sdg}[~\num{11} and \num{13}]~\cite{unitednationsdepartmentofeconomicandsocialaffairsGoal11,unitednationsdepartmentofeconomicandsocialaffairsGoal13}, of which sustainable transport is a subtheme of \gls{sdg}[~\num{11}]~\cite{unitednationsdepartmentofeconomicandsocialaffairsSustainableTransport}. These two \glspl{sdg} share common goals. \Pgls{un} note that the transport sector will play a particularly important role in achieving the Paris Agreement, as ``nearly \SI{25}{\percent} of energy-related greenhouse gases come from transport'' and these emissions are projected to grow substantially, especially in lower income countries~\cite{unitednationsdepartmentofeconomicandsocialaffairsSustainableTransport, intergovernmentalpanelonclimatechangeipccTransport2023}. Many countries have set specific transport-related targets to meet these broader net-zero and decarbonisation goals: for example, \num{68} countries have set a target of \SI{100}{\percent} sales of \glspl{ev} at a future point~\cite{partnershiponsustainablelowcarbontransportTransportClimateSustainability2025}.

\begin{figure*}[ht!]
    \centering
    \includegraphics[width=\linewidth]{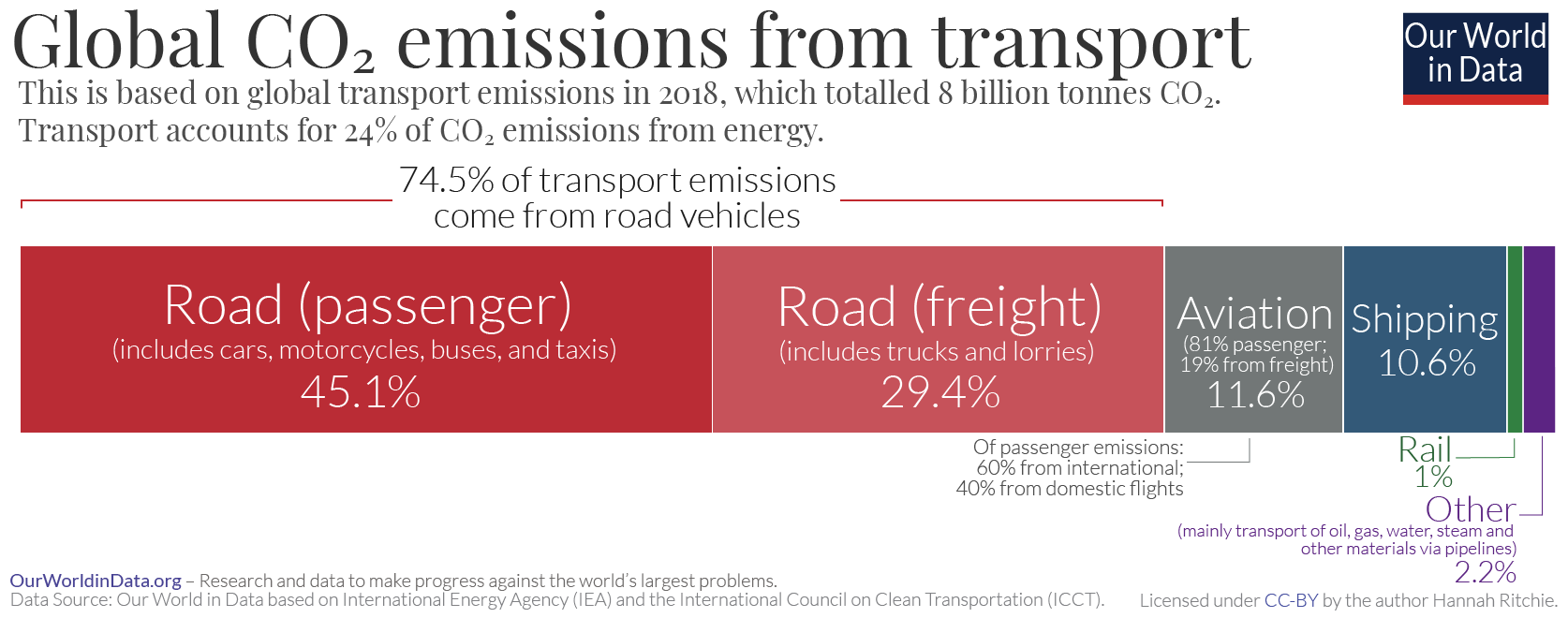}
    \caption{Global \ce{CO_2} emissions from transport. Image courtesy of Hannah Ritchie \cite{ritchieCarsPlanesTrains2020} and licensed for redistribution under CC BY.}
    \label{fig:emissions}
\end{figure*}

However, progress and impact towards transport decarbonisation remains split across different modes. \Cref{fig:emissions} shows that both road passenger and freight transport contribute \SI{74.5}{\percent} of total transport \ce{CO_2} emissions~\cite{ritchieCarsPlanesTrains2020}. Despite recent rising sales of \glspl{ev} (\SI{25}{\percent} in 2024), they only constitute \SI{5}{\percent} of cars worldwide~\cite{partnershiponsustainablelowcarbontransportTransportClimateSustainability2025}. In contrast, rail produces just \SI{1}{\percent} of global \ce{CO2} emissions~\cite{ritchieCarsPlanesTrains2020}, and gets \SI{15}{\percent} of its energy from renewable sources~\cite{partnershiponsustainablelowcarbontransportTransportClimateSustainability2025}, yet accounts for \SI{12}{\percent} of global passenger transport~\cite{partnershiponsustainablelowcarbontransportTransportClimateSustainability2025}.

This disparity reflects the diversity of the transportation sector and the societies it seeks to provide services for. Both are complex systems, where successful decarbonisation strategies will rely on a multitude of quantitative and qualitative techniques working in tandem across a number of transport modes. 

Key to enabling this holistic approach is a means of sharing data that is able to link across transport modes, communicate different data sources, and capture different data models. However, as reflected by the wider sector, many data sources remain siloed and incompatible with one another, and lack a semantic layer that provide descriptions of the data, making it challenging to share and integrate. This challenge is known as \emph{semantic heterogeneity} and is pervasive both within transport and across other domains~\cite{shethFederatedDatabaseSystems1990, scroccaIntelligentUrbanTraffic2025}.

A well-established approach to addressing semantic heterogeneity is the use of \textit{semantic technologies}. This includes ontologies, which provide explicit, machine-interpretable specifications of the concepts, relationships, and logical axioms within a domain~\cite{staabHandbookOntologies2013}, and knowledge graphs, that represent data as a network of real-world entities (nodes) and the relationships between them (edges).

A subset of semantic technologies is \textit{Semantic Web technologies}, which relates specifically to the standards maintained by the \gls{w3c}. Within the Semantic Web technology stack, \gls{rdf}~\cite{worldwidewebconsortiumRDFSemanticWeb2014} and \gls{uri}[--based] identification provide a basis for linking and publishing data, while \pgls{owl}~\cite{grauOWL2Next2008,worldwidewebconsortiumOWL2Web2012} provides formal semantics to create ontologies in \gls{rdf} and supports automated reasoning to infer new facts from existing information based on \glspl{dl}~\cite{baaderIntroductionDescriptionLogic2017}. \Gls{sparql}~\cite{worldwidewebconsortiumSPARQL11Query2013} provides a way to query against data stored in the \gls{rdf} format. Providing explicit vocabularies, formal semantics, and reusable identifiers as Semantic Web standards contributes to \gls{fair} data practices~\cite{wilkinsonFAIRGuidingPrinciples2016}, and promotes interoperability, portability, and long-term reuse. This is crucial for a sector as large and multidisciplinary as transport.

As many future transport systems will be data-driven, relying on the integration of information from vehicles, infrastructure, public transport operators, logistics providers, energy networks, and policymakers is a growing challenge. Adopting Semantic Web technologies as a foundational enabling technology can support the high-level objectives of transport decarbonisation by providing a common framework for data description, sharing, and inference, directly addressing semantic heterogeneity, turning fragmented data into actionable knowledge.  

\subsection{Research Questions and Contributions}
The motivation of this study is to understand the current state-of-the-art in the development and use of Semantic Web technologies in supporting transport decarbonisation. Previous work has reviewed the intersection of transport and ontologies~\cite{katsumiOntologiesTransportationResearch2018} and the use of \glspl{llm} for transport knowledge graph construction~\cite{lingReviewKnowledgeGraph2026}. However, to the best of our knowledge, no previous work has surveyed the intersection of transport and decarbonisation with Semantic Web technologies. Therefore we conduct a scoping review, which seeks to determine how much literature on this topic exists and its characteristics~\cite{boothSystematicApproachesSuccessful2022}. Specifically, we answer the following research questions:
\begin{enumerate}
    \item \label{rq1} What (if any) aspects of semantic technologies are used to address transport decarbonisation?
    \item \label{rq2} Which studies adopt Semantic Web technologies?
    \item \label{rq3} Do studies provide publicly available artefacts, and are they aligned with \gls{w3c} standards and the \gls{fair} principles?
    \item \label{rq4} What are the emerging research trends within the existing literature?
    \item \label{rq5} What are the directions for future research at this intersection?
\end{enumerate}

We conducted a structured search of the Scopus database for the period of 2007--2026, using a search string that combines terms for semantic technologies, transport, and decarbonisation. After filtering by Scopus indexed keywords from an original corpus of \num{222}, we retained \num{60} studies for in‑depth analysis.

Our study makes three main contributions to the literature:
\begin{enumerate}
    \item First, we map the literature and find that only seven studies address all three fields simultaneously. This reveals that the potential of Semantic Web technologies for transport decarbonisation remains largely unrealised.
    \item Second, we cross-check the literature against a set of definitions around semantic technologies, and find the majority of studies are adhering to \gls{w3c} standards. In addition, our review identifies that a minority of studies provide publicly accessible artefacts, hindering the \gls{fair} principles.
    \item Third, we highlight critical gaps. Decarbonisation is rarely treated as a \textit{queryable} concept; emissions, when present, are typically implicit or aggregated. Consequently, leveraging semantic technologies for reasoning on low‑carbon operational decisions remains largely unsupported by existing work.
\end{enumerate}

\subsection{Paper Structure}

\Cref{sec:background} provides background on Semantic Web technologies using a transport use-case. \Cref{sec:methods} describes our search to obtain the initial corpus of \num{222} studies, and the filtering methodology to reduce it to \num{60} relevant studies. \Cref{sec:results} carries out the methods and presents the findings of the \num{60} key studies identified, answering \Cref{rq1,rq2,rq3}. \Cref{sec:discussion} discusses key trends, terminological drift, and potential future research, answering \Cref{rq5,rq5}. \Cref{sec:conclusion} concludes the work, and provides a general overview of the findings of the paper.

\section{Background: A Semantic Web Technologies Workflow} \label{sec:background}

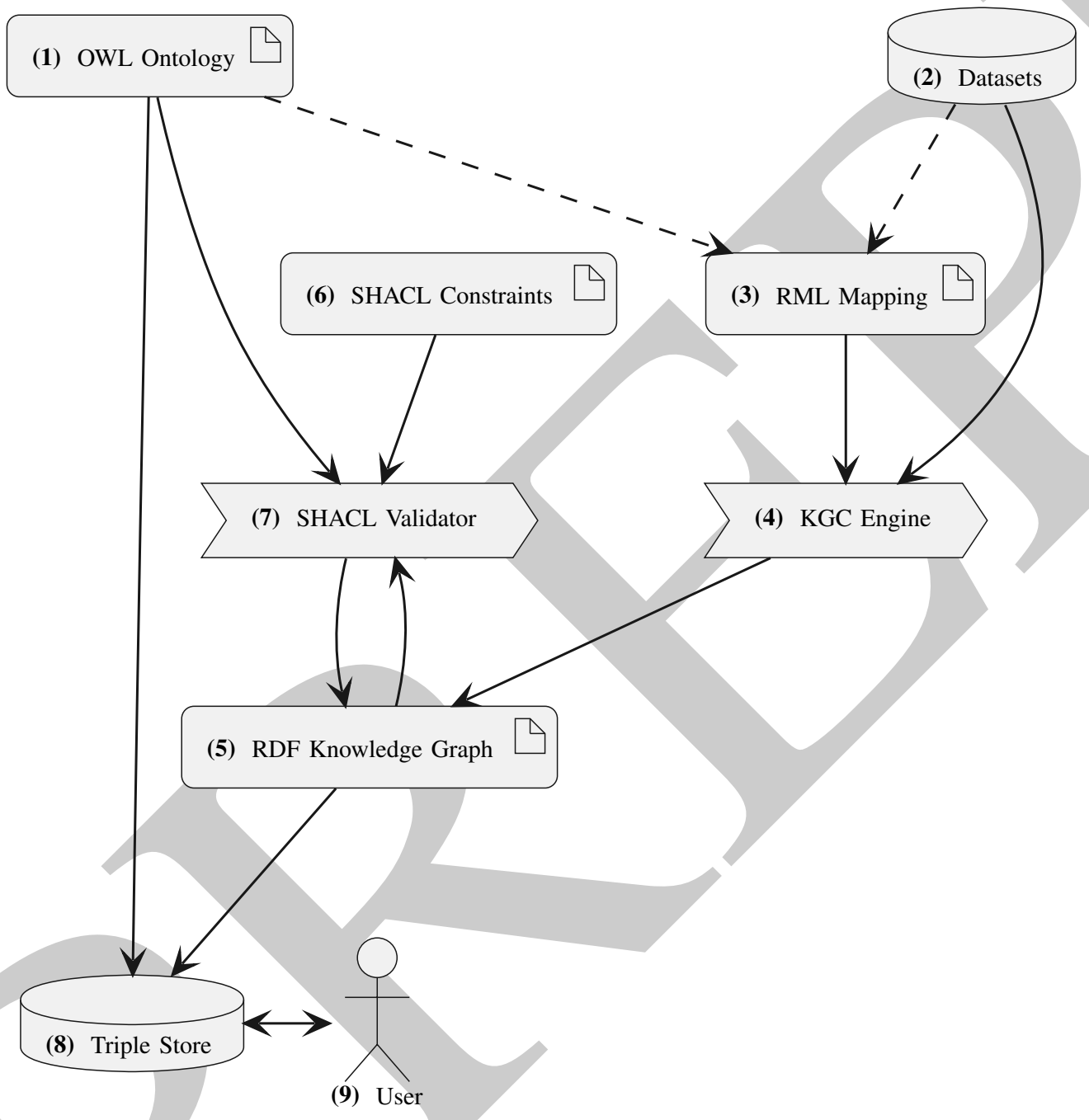
\begin{figure}[ht!]
    \centering
    \resizebox{\linewidth}{!}{%
        \definecolor{plantucolor0000}{RGB}{241,241,241}
        \definecolor{plantucolor0001}{RGB}{24,24,24}
        \definecolor{plantucolor0002}{RGB}{0,0,0}
        \begin{tikzpicture}[yscale=-1
        ,pstyle0/.style={color=plantucolor0001,fill=plantucolor0000,line width=0.5pt}
        ,pstyle1/.style={color=plantucolor0001,line width=0.5pt}
        ,pstyle2/.style={color=plantucolor0001,line width=1.0pt,dash pattern=on 7.0pt off 7.0pt}
        ,pstyle3/.style={color=plantucolor0001,fill=plantucolor0001,line width=1.0pt}
        ,pstyle4/.style={color=plantucolor0001,line width=1.0pt}
        ]
        \draw[pstyle0] (288.5pt,198pt) -- (392.57pt,198pt) -- (402.57pt,213pt) -- (392.57pt,228pt) -- (288.5pt,228pt) -- (298.5pt,213pt) -- cycle;
        \node at (308.5pt,208pt)[below right,color=black,inner sep=0]{\textbf{(4)}};
        \node at (323.19pt,208.45pt)[below right,color=black,inner sep=0]{~KGC Engine};
        \draw[pstyle0] (85.5pt,198pt) -- (211.2pt,198pt) -- (221.2pt,213pt) -- (211.2pt,228pt) -- (85.5pt,228pt) -- (95.5pt,213pt) -- cycle;
        \node at (105.5pt,208pt)[below right,color=black,inner sep=0]{\textbf{(7)}};
        \node at (120.19pt,208.34pt)[below right,color=black,inner sep=0]{~SHACL Validator};
        \draw[pstyle0] (7pt,14pt) arc (180:270:5pt) -- (12pt,9pt) -- (117.39pt,9pt) arc (270:360:5pt) -- (122.39pt,14pt) -- (122.39pt,37pt) arc (0:90:5pt) -- (117.39pt,42pt) -- (12pt,42pt) arc (90:180:5pt) -- (7pt,37pt) -- cycle;
        \draw[pstyle0] (105.39pt,14pt) -- (105.39pt,28pt) -- (117.39pt,28pt) -- (117.39pt,20pt) -- (111.39pt,14pt) -- (105.39pt,14pt) -- cycle;
        \draw[pstyle1] (111.39pt,14pt) -- (111.39pt,20pt);
        \draw[pstyle1] (117.39pt,20pt) -- (111.39pt,20pt);
        \node at (17pt,22pt)[below right,color=black,inner sep=0]{\textbf{(1)}};
        \node at (31.69pt,22.45pt)[below right,color=black,inner sep=0]{~OWL Ontology};
        \draw[pstyle0] (289pt,110pt) arc (180:270:5pt) -- (294pt,105pt) -- (396.76pt,105pt) arc (270:360:5pt) -- (401.76pt,110pt) -- (401.76pt,133pt) arc (0:90:5pt) -- (396.76pt,138pt) -- (294pt,138pt) arc (90:180:5pt) -- (289pt,133pt) -- cycle;
        \draw[pstyle0] (384.76pt,110pt) -- (384.76pt,124pt) -- (396.76pt,124pt) -- (396.76pt,116pt) -- (390.76pt,110pt) -- (384.76pt,110pt) -- cycle;
        \draw[pstyle1] (390.76pt,110pt) -- (390.76pt,116pt);
        \draw[pstyle1] (396.76pt,116pt) -- (390.76pt,116pt);
        \node at (299pt,118pt)[below right,color=black,inner sep=0]{\textbf{(3)}};
        \node at (313.69pt,118.67pt)[below right,color=black,inner sep=0]{~RML Mapping};
        \draw[pstyle0] (117.5pt,110pt) arc (180:270:5pt) -- (122.5pt,105pt) -- (248.02pt,105pt) arc (270:360:5pt) -- (253.02pt,110pt) -- (253.02pt,133pt) arc (0:90:5pt) -- (248.02pt,138pt) -- (122.5pt,138pt) arc (90:180:5pt) -- (117.5pt,133pt) -- cycle;
        \draw[pstyle0] (236.02pt,110pt) -- (236.02pt,124pt) -- (248.02pt,124pt) -- (248.02pt,116pt) -- (242.02pt,110pt) -- (236.02pt,110pt) -- cycle;
        \draw[pstyle1] (242.02pt,110pt) -- (242.02pt,116pt);
        \draw[pstyle1] (248.02pt,116pt) -- (242.02pt,116pt);
        \node at (127.5pt,118pt)[below right,color=black,inner sep=0]{\textbf{(6)}};
        \node at (142.19pt,118.34pt)[below right,color=black,inner sep=0]{~SHACL Constraints};
        \draw[pstyle0] (362.5pt,16pt) ..controls (362.5pt,6pt) and (400.38pt,6pt) .. (400.38pt,6pt) ..controls (400.38pt,6pt) and (438.26pt,6pt) .. (438.26pt,16pt) -- (438.26pt,35pt) ..controls (438.26pt,45pt) and (400.38pt,45pt) .. (400.38pt,45pt) ..controls (400.38pt,45pt) and (362.5pt,45pt) .. (362.5pt,35pt) -- (362.5pt,16pt);
        \draw[pstyle1] (362.5pt,16pt) ..controls (362.5pt,26pt) and (400.38pt,26pt) .. (400.38pt,26pt) ..controls (400.38pt,26pt) and (438.26pt,26pt) .. (438.26pt,16pt);
        \node at (372.5pt,30pt)[below right,color=black,inner sep=0]{\textbf{(2)}};
        \node at (387.19pt,30.67pt)[below right,color=black,inner sep=0]{~Datasets};
        \draw[pstyle0] (77.5pt,293pt) arc (180:270:5pt) -- (82.5pt,288pt) -- (224.1pt,288pt) arc (270:360:5pt) -- (229.1pt,293pt) -- (229.1pt,316pt) arc (0:90:5pt) -- (224.1pt,321pt) -- (82.5pt,321pt) arc (90:180:5pt) -- (77.5pt,316pt) -- cycle;
        \draw[pstyle0] (212.1pt,293pt) -- (212.1pt,307pt) -- (224.1pt,307pt) -- (224.1pt,299pt) -- (218.1pt,293pt) -- (212.1pt,293pt) -- cycle;
        \draw[pstyle1] (218.1pt,293pt) -- (218.1pt,299pt);
        \draw[pstyle1] (224.1pt,299pt) -- (218.1pt,299pt);
        \node at (87.5pt,301pt)[below right,color=black,inner sep=0]{\textbf{(5)}};
        \node at (102.19pt,301.45pt)[below right,color=black,inner sep=0]{~RDF Knowledge Graph};
        \draw[pstyle0] (12.5pt,406.5pt) ..controls (12.5pt,396.5pt) and (57.515pt,396.5pt) .. (57.515pt,396.5pt) ..controls (57.515pt,396.5pt) and (102.53pt,396.5pt) .. (102.53pt,406.5pt) -- (102.53pt,425.5pt) ..controls (102.53pt,435.5pt) and (57.515pt,435.5pt) .. (57.515pt,435.5pt) ..controls (57.515pt,435.5pt) and (12.5pt,435.5pt) .. (12.5pt,425.5pt) -- (12.5pt,406.5pt);
        \draw[pstyle1] (12.5pt,406.5pt) ..controls (12.5pt,416.5pt) and (57.515pt,416.5pt) .. (57.515pt,416.5pt) ..controls (57.515pt,416.5pt) and (102.53pt,416.5pt) .. (102.53pt,406.5pt);
        \node at (22.5pt,420.5pt)[below right,color=black,inner sep=0]{\textbf{(8)}};
        \node at (37.19pt,420.95pt)[below right,color=black,inner sep=0]{~Triple Store};
        \draw[pstyle0] (156.41pt,389.5pt) ellipse (8pt and 8pt);
        \draw[pstyle1] (156.41pt,397.5pt) -- (156.41pt,424.5pt)(143.41pt,405.5pt) -- (169.41pt,405.5pt)(156.41pt,424.5pt) -- (143.41pt,439.5pt)(156.41pt,424.5pt) -- (169.41pt,439.5pt);
        \node at (137.5pt,441pt)[below right,color=black,inner sep=0]{\textbf{(9)}};
        \node at (152.19pt,441.67pt)[below right,color=black,inner sep=0]{~User};
        \draw[pstyle2] (110.9pt,42.02pt) ..controls (162.97pt,59.44pt) and (241.2498pt,85.6271pt) .. (293.3398pt,103.0471pt);
        \draw[pstyle3] (299.03pt,104.95pt) -- (291.7633pt,98.3021pt) -- (294.2881pt,103.3642pt) -- (289.226pt,105.8891pt) -- (299.03pt,104.95pt) -- cycle;
        \draw[pstyle2] (389.63pt,45.07pt) ..controls (379.43pt,62.51pt) and (367.3612pt,83.142pt) .. (357.7312pt,99.592pt);
        \draw[pstyle3] (354.7pt,104.77pt) -- (362.6989pt,99.0239pt) -- (357.226pt,100.455pt) -- (355.7949pt,94.9822pt) -- (354.7pt,104.77pt) -- cycle;
        \draw[pstyle4] (345.5pt,138.09pt) ..controls (345.5pt,154.92pt) and (345.5pt,175.49pt) .. (345.5pt,191.77pt);
        \draw[pstyle3] (345.5pt,197.77pt) -- (349.5pt,188.77pt) -- (345.5pt,192.77pt) -- (341.5pt,188.77pt) -- (345.5pt,197.77pt) -- cycle;
        \draw[pstyle4] (409.82pt,45.35pt) ..controls (419.7pt,68.11pt) and (432.27pt,107.16pt) .. (419.5pt,138pt) ..controls (408.93pt,163.51pt) and (389.6671pt,181.2701pt) .. (371.8171pt,194.3701pt);
        \draw[pstyle3] (366.98pt,197.92pt) -- (376.6023pt,195.8198pt) -- (371.0109pt,194.9617pt) -- (371.8691pt,189.3703pt) -- (366.98pt,197.92pt) -- cycle;
        \draw[pstyle4] (315.08pt,228.18pt) ..controls (280.18pt,244.45pt) and (228.588pt,268.4946pt) .. (192.448pt,285.3446pt);
        \draw[pstyle3] (187.01pt,287.88pt) -- (196.8573pt,287.7022pt) -- (191.5417pt,285.7672pt) -- (193.4767pt,280.4516pt) -- (187.01pt,287.88pt) -- cycle;
        \draw[pstyle4] (179.93pt,138.09pt) ..controls (173.91pt,154.92pt) and (166.4298pt,175.8402pt) .. (160.6098pt,192.1202pt);
        \draw[pstyle3] (158.59pt,197.77pt) -- (165.3862pt,190.6418pt) -- (160.2731pt,193.0618pt) -- (157.8531pt,187.9487pt) -- (158.59pt,197.77pt) -- cycle;
        \draw[pstyle4] (164.02pt,287.95pt) ..controls (168.09pt,271.18pt) and (169.5063pt,250.4952pt) .. (165.2463pt,234.1852pt);
        \draw[pstyle3] (163.73pt,228.38pt) -- (162.1342pt,238.0987pt) -- (164.9936pt,233.2177pt) -- (169.8746pt,236.077pt) -- (163.73pt,228.38pt) -- cycle;
        \draw[pstyle4] (143.84pt,228.38pt) ..controls (139.82pt,244.69pt) and (138.3808pt,265.3314pt) .. (142.2208pt,282.1014pt);
        \draw[pstyle3] (143.56pt,287.95pt) -- (145.4503pt,278.2842pt) -- (142.444pt,283.0761pt) -- (137.6521pt,280.0699pt) -- (143.56pt,287.95pt) -- cycle;
        \draw[pstyle4] (67.74pt,42.27pt) ..controls (72.67pt,64.45pt) and (83.23pt,105.76pt) .. (99.5pt,138pt) ..controls (110.89pt,160.58pt) and (125.0839pt,179.07pt) .. (137.1439pt,193.32pt);
        \draw[pstyle3] (141.02pt,197.9pt) -- (138.2592pt,188.446pt) -- (137.7899pt,194.0834pt) -- (132.1526pt,193.6141pt) -- (141.02pt,197.9pt) -- cycle;
        \draw[pstyle4] (64.18pt,42.12pt) ..controls (63.36pt,82.68pt) and (61.13pt,194.7pt) .. (59.5pt,288pt) ..controls (58.83pt,326.58pt) and (58.2298pt,365.7607pt) .. (57.8598pt,390.4907pt);
        \draw[pstyle3] (57.77pt,396.49pt) -- (61.9042pt,387.5508pt) -- (57.8448pt,391.4906pt) -- (53.9051pt,387.4312pt) -- (57.77pt,396.49pt) -- cycle;
        \draw[pstyle4] (139.75pt,321.19pt) ..controls (122.33pt,341.06pt) and (96.3053pt,370.7382pt) .. (77.7553pt,391.8982pt);
        \draw[pstyle3] (73.8pt,396.41pt) -- (82.7407pt,392.2792pt) -- (77.096pt,392.6502pt) -- (76.725pt,387.0055pt) -- (73.8pt,396.41pt) -- cycle;
        \draw[pstyle4] (108.75pt,416pt) ..controls (120.22pt,416pt) and (119.7pt,416pt) .. (131.18pt,416pt);
        \draw[pstyle3] (102.75pt,416pt) -- (111.75pt,420pt) -- (107.75pt,416pt) -- (111.75pt,412pt) -- (102.75pt,416pt) -- cycle;
        \draw[pstyle3] (137.18pt,416pt) -- (128.18pt,412pt) -- (132.18pt,416pt) -- (128.18pt,420pt) -- (137.18pt,416pt) -- cycle;
        \end{tikzpicture}
    }
    \caption{A typical workflow using Semantic Web technologies.}
    \label{fig:example-workflow}
\end{figure}

\Cref{fig:example-workflow} illustrates a typical workflow using \gls{w3c}[-endorsed] Semantic Web technologies. 

\begin{itemize}
    \item It begins with the development of an \gls{owl} ontology \textbf{(1)}, which provides a machine-readable conceptualisation of the relevant domain. In a transport setting, such an ontology may describe classes such as \texttt{Vehicle}, \texttt{Route}, \texttt{ChargingStation}, or \texttt{EmissionFactor}; object properties such as \texttt{servesRoute}, \texttt{locatedAt}, or \texttt{usesEnergySource}; data properties such as \texttt{batteryCapacity} or \texttt{CO2ePerKm}; and individuals representing concrete assets, services, or locations (such as an actual instance of a \texttt{Vehicle}). Each entity is identified using an \gls{iri}, allowing data from different sources to refer to the same underlying concepts in a consistent and reusable way.
    \item Once an ontology has been defined, heterogeneous datasets \textbf{(2)} can be mapped to it using a mapping language such as \gls{r2rml}~\cite{worldwidewebconsortiumR2RMLRDBRDF2012} or \gls{rml}~\cite{dimouRMLGenericLanguage2014}. These mappings \textbf{(3)} specify how fields in source data correspond to ontological entities and properties. For example, a \gls{csv} column containing charger power ratings, an \gls{api} field describing station locations, and a database table listing vehicle identifiers can all be mapped into the common semantic representation provided by the \gls{owl} ontology. 
    \item \Gls{kgc} engines \textbf{(4)} such as \textit{Morph-KGC}~\cite{arenas-guerreroMorphKGCScalableKnowledge2024} can then execute these mappings to generate an \gls{rdf} knowledge graph \textbf{(5)} that contains the heterogeneous source data in a single graph-based representation.
    \item The resulting \gls{rdf} knowledge graph can then be validated using \gls{shacl} shapes \textbf{(6)} and a \gls{shacl} validator \textbf{(7)}~\cite{worldwidewebconsortiumShapesConstraintLanguage2017}, which provides constraints over graph data. For example, this can be used to check that every charging station has a location, every trip is associated with an operator, or every emissions record provides values in the expected datatype. Validation is particularly important when integrating data from multiple sources, as it helps detect inconsistencies, missing values, and schema violations before querying or analysis of the data.
    \item After validation, the \gls{rdf} knowledge graph can be loaded into a triple store \textbf{(8)} and, where appropriate, combined with a reasoner to infer new facts about the existing data. For example, if a vehicle has a battery-electric drivetrain and is assigned to a route served by specific charging infrastructure, additional inferences may classify it as part of an specific electrified service system, even when that fact was not explicitly stated in the original data. A user \textbf{(9)} can then use \gls{sparql} to query the data, including these newly inferred facts.
\end{itemize}

\section{Scoping Review Methodology} \label{sec:methods}

This section describes the methodology to collect, categorise, and analyse the literature. We followed a structured search strategy to identify relevant publications around \textit{semantic technologies}, \textit{transport}, and \textit{decarbonisation}: henceforth, these are collectively referred to as the three \textit{thematic pillars}.

\begin{figure}[ht]
    \centering
    \resizebox{\linewidth}{!}{%
        \definecolor{plantucolor0000}{RGB}{34,34,34}
        \definecolor{plantucolor0001}{RGB}{241,241,241}
        \definecolor{plantucolor0002}{RGB}{24,24,24}
        \definecolor{plantucolor0003}{RGB}{0,0,0}
        \definecolor{plantucolor0004}{RGB}{85,85,85}
        \begin{tikzpicture}[yscale=-1
        ,pstyle0/.style={color=plantucolor0000,fill=plantucolor0000,line width=1.0pt}
        ,pstyle1/.style={color=plantucolor0002,fill=plantucolor0001,line width=0.5pt}
        ,pstyle2/.style={color=plantucolor0004,fill=plantucolor0004,line width=1.0pt}
        ,pstyle4/.style={color=plantucolor0002,line width=1.0pt}
        ,pstyle5/.style={color=plantucolor0002,fill=plantucolor0002,line width=1.0pt}
        ]
        \draw[pstyle0] (211.885pt,20pt) ellipse (10pt and 10pt);
        \draw[pstyle1] (152.73pt,65pt) arc (180:270:15pt) -- (167.73pt,50pt) -- (256.04pt,50pt) arc (270:360:15pt) -- (271.04pt,65pt) -- (271.04pt,65pt) arc (0:90:15pt) -- (256.04pt,80pt) -- (167.73pt,80pt) arc (90:180:15pt) -- (152.73pt,65pt) -- cycle;
        \node at (162.73pt,61.475pt)[below right,color=black,inner sep=0]{Generate search string};
        \draw[pstyle1] (166.515pt,115pt) arc (180:270:15pt) -- (181.515pt,100pt) -- (242.255pt,100pt) arc (270:360:15pt) -- (257.255pt,115pt) -- (257.255pt,115pt) arc (0:90:15pt) -- (242.255pt,130pt) -- (181.515pt,130pt) arc (90:180:15pt) -- (166.515pt,115pt) -- cycle;
        \node at (176.515pt,111.475pt)[below right,color=black,inner sep=0]{Query SCOPUS};
        \draw[pstyle2] (17.695pt,153pt) arc (180:270:3pt) -- (20.695pt,150pt) -- (403.075pt,150pt) arc (270:360:3pt) -- (406.075pt,153pt) -- (406.075pt,153pt) arc (0:90:3pt) -- (403.075pt,156pt) -- (20.695pt,156pt) arc (90:180:3pt) -- (17.695pt,153pt) -- cycle;
        \draw[pstyle1] (31.695pt,223.5pt) arc (180:270:20pt) -- (51.695pt,203.5pt) -- (96.675pt,203.5pt) arc (270:360:20pt) -- (116.675pt,223.5pt) -- (116.675pt,223.5pt) arc (0:90:20pt) -- (96.675pt,243.5pt) -- (51.695pt,243.5pt) arc (90:180:20pt) -- (31.695pt,223.5pt) -- cycle;
        \node at (41.695pt,215.1pt)[below right,color=black,inner sep=0]{Extract corpus};
        \node at (41.695pt,225.17pt)[below right,color=black,inner sep=0]{\textit{n=222}};
        \draw[pstyle1] (198.175pt,191pt) arc (180:270:15pt) -- (213.175pt,176pt) -- (323.575pt,176pt) arc (270:360:15pt) -- (338.575pt,191pt) -- (338.575pt,191pt) arc (0:90:15pt) -- (323.575pt,206pt) -- (213.175pt,206pt) arc (90:180:15pt) -- (198.175pt,191pt) -- cycle;
        \node at (208.175pt,187.475pt)[below right,color=black,inner sep=0]{Extract SCOPUS keywords};
        \draw[pstyle1] (144.675pt,256pt) arc (180:270:15pt) -- (159.675pt,241pt) -- (377.075pt,241pt) arc (270:360:15pt) -- (392.075pt,256pt) -- (392.075pt,256pt) arc (0:90:15pt) -- (377.075pt,271pt) -- (159.675pt,271pt) arc (90:180:15pt) -- (144.675pt,256pt) -- cycle;
        \node at (154.675pt,251pt)[below right,color=black,inner sep=0]{Select relevant indexed keywords per Pillar~};
        \node at (346.645pt,251.3pt)[below right,color=black,inner sep=0]{\textbf{T}};
        \node at (354.645pt,256.99pt)[below right,color=black,inner sep=0]{,~};
        \node at (360.755pt,251.19pt)[below right,color=black,inner sep=0]{\textbf{D}};
        \node at (369.575pt,256.99pt)[below right,color=black,inner sep=0]{,~};
        \node at (375.685pt,251.08pt)[below right,color=black,inner sep=0]{\textbf{S}};
        \draw[pstyle2] (17.695pt,294pt) arc (180:270:3pt) -- (20.695pt,291pt) -- (403.075pt,291pt) arc (270:360:3pt) -- (406.075pt,294pt) -- (406.075pt,294pt) arc (0:90:3pt) -- (403.075pt,297pt) -- (20.695pt,297pt) arc (90:180:3pt) -- (17.695pt,294pt) -- cycle;
        \draw[pstyle1] (96.665pt,337.11pt) arc (180:270:20.11pt) -- (116.775pt,317pt) -- (306.995pt,317pt) arc (270:360:20.11pt) -- (327.105pt,337.11pt) -- (327.105pt,337.11pt) arc (0:90:20.11pt) -- (306.995pt,357.22pt) -- (116.775pt,357.22pt) arc (90:180:20.11pt) -- (96.665pt,337.11pt) -- cycle;
        \node at (106.665pt,328.42pt)[below right,color=black,inner sep=0]{Assign papers to relevant pillars using keywords};
        \node at (106.665pt,338.89pt)[below right,color=black,inner sep=0]{\textit{n=222}};
        \draw[pstyle1] (91.745pt,397.275pt) arc (180:270:20.055pt) -- (111.8pt,377.22pt) -- (311.97pt,377.22pt) arc (270:360:20.055pt) -- (332.025pt,397.275pt) -- (332.025pt,397.275pt) arc (0:90:20.055pt) -- (311.97pt,417.33pt) -- (111.8pt,417.33pt) arc (90:180:20.055pt) -- (91.745pt,397.275pt) -- cycle;
        \node at (101.745pt,388.695pt)[below right,color=black,inner sep=0]{Identify relevant regions based on the three pillars};
        \node at (101.745pt,399pt)[below right,color=black,inner sep=0]{\textit{n=60}};
        \draw[pstyle2] (11pt,440.33pt) arc (180:270:3pt) -- (14pt,437.33pt) -- (409.77pt,437.33pt) arc (270:360:3pt) -- (412.77pt,440.33pt) -- (412.77pt,440.33pt) arc (0:90:3pt) -- (409.77pt,443.33pt) -- (14pt,443.33pt) arc (90:180:3pt) -- (11pt,440.33pt) -- cycle;
        \draw[pstyle1] (25pt,483.33pt) arc (180:270:20pt) -- (45pt,463.33pt) -- (107.71pt,463.33pt) arc (270:360:20pt) -- (127.71pt,483.33pt) -- (127.71pt,483.33pt) arc (0:90:20pt) -- (107.71pt,503.33pt) -- (45pt,503.33pt) arc (90:180:20pt) -- (25pt,483.33pt) -- cycle;
        \node at (35pt,473.33pt)[below right,color=black,inner sep=0]{Review Region~};
        \node at (103.32pt,473.19pt)[below right,color=black,inner sep=0]{\textbf{TS}};
        \node at (35pt,485pt)[below right,color=black,inner sep=0]{\textit{n=50}};
        \draw[pstyle1] (155.71pt,483.33pt) arc (180:270:20pt) -- (175.71pt,463.33pt) -- (239.24pt,463.33pt) arc (270:360:20pt) -- (259.24pt,483.33pt) -- (259.24pt,483.33pt) arc (0:90:20pt) -- (239.24pt,503.33pt) -- (175.71pt,503.33pt) arc (90:180:20pt) -- (155.71pt,483.33pt) -- cycle;
        \node at (165.71pt,473.33pt)[below right,color=black,inner sep=0]{Review Region~};
        \node at (234.03pt,473.19pt)[below right,color=black,inner sep=0]{\textbf{DS}};
        \node at (165.71pt,485pt)[below right,color=black,inner sep=0]{\textit{n=3}};
        \draw[pstyle1] (287.24pt,483.33pt) arc (180:270:20pt) -- (307.24pt,463.33pt) -- (378.77pt,463.33pt) arc (270:360:20pt) -- (398.77pt,483.33pt) -- (398.77pt,483.33pt) arc (0:90:20pt) -- (378.77pt,503.33pt) -- (307.24pt,503.33pt) arc (90:180:20pt) -- (287.24pt,483.33pt) -- cycle;
        \node at (297.24pt,473.33pt)[below right,color=black,inner sep=0]{Review Region~};
        \node at (365.56pt,473.19pt)[below right,color=black,inner sep=0]{\textbf{TDS}};
        \node at (297.24pt,485pt)[below right,color=black,inner sep=0]{\textit{n=7}};
        \draw[pstyle1] (211.885pt,523.33pt) -- (223.885pt,535.33pt) -- (211.885pt,547.33pt) -- (199.885pt,535.33pt) -- (211.885pt,523.33pt) -- cycle;
        \draw[pstyle1] (136.625pt,587.385pt) arc (180:270:20.055pt) -- (156.68pt,567.33pt) -- (267.09pt,567.33pt) arc (270:360:20.055pt) -- (287.145pt,587.385pt) -- (287.145pt,587.385pt) arc (0:90:20.055pt) -- (267.09pt,607.4399pt) -- (156.68pt,607.4399pt) arc (90:180:20.055pt) -- (136.625pt,587.385pt) -- cycle;
        \node at (146.625pt,578.805pt)[below right,color=black,inner sep=0]{Crosscheck against definitions};
        \node at (146.625pt,589.1099pt)[below right,color=black,inner sep=0]{\textit{n=60}};
        \draw[color=plantucolor0000,line width=1.0pt] (211.885pt,638.4399pt) ellipse (11pt and 11pt);
        \draw[pstyle0] (211.885pt,638.4399pt) ellipse (6pt and 6pt);
        \draw[pstyle4] (211.885pt,30pt) -- (211.885pt,50pt);
        \draw[pstyle5] (207.885pt,40pt) -- (211.885pt,50pt) -- (215.885pt,40pt) -- (211.885pt,44pt) -- cycle;
        \draw[pstyle4] (211.885pt,80pt) -- (211.885pt,100pt);
        \draw[pstyle5] (207.885pt,90pt) -- (211.885pt,100pt) -- (215.885pt,90pt) -- (211.885pt,94pt) -- cycle;
        \draw[pstyle4] (268.375pt,206pt) -- (268.375pt,241pt);
        \draw[pstyle5] (264.375pt,231pt) -- (268.375pt,241pt) -- (272.375pt,231pt) -- (268.375pt,235pt) -- cycle;
        \draw[pstyle4] (74.185pt,156pt) -- (74.185pt,203.5pt);
        \draw[pstyle5] (70.185pt,193.5pt) -- (74.185pt,203.5pt) -- (78.185pt,193.5pt) -- (74.185pt,197.5pt) -- cycle;
        \draw[pstyle4] (268.375pt,156pt) -- (268.375pt,176pt);
        \draw[pstyle5] (264.375pt,166pt) -- (268.375pt,176pt) -- (272.375pt,166pt) -- (268.375pt,170pt) -- cycle;
        \draw[pstyle4] (74.185pt,243.5pt) -- (74.185pt,291pt);
        \draw[pstyle5] (70.185pt,281pt) -- (74.185pt,291pt) -- (78.185pt,281pt) -- (74.185pt,285pt) -- cycle;
        \draw[pstyle4] (268.375pt,271pt) -- (268.375pt,291pt);
        \draw[pstyle5] (264.375pt,281pt) -- (268.375pt,291pt) -- (272.375pt,281pt) -- (268.375pt,285pt) -- cycle;
        \draw[pstyle4] (211.885pt,130pt) -- (211.885pt,150pt);
        \draw[pstyle5] (207.885pt,140pt) -- (211.885pt,150pt) -- (215.885pt,140pt) -- (211.885pt,144pt) -- cycle;
        \draw[pstyle4] (211.885pt,297pt) -- (211.885pt,317pt);
        \draw[pstyle5] (207.885pt,307pt) -- (211.885pt,317pt) -- (215.885pt,307pt) -- (211.885pt,311pt) -- cycle;
        \draw[pstyle4] (211.885pt,357.22pt) -- (211.885pt,377.22pt);
        \draw[pstyle5] (207.885pt,367.22pt) -- (211.885pt,377.22pt) -- (215.885pt,367.22pt) -- (211.885pt,371.22pt) -- cycle;
        \draw[pstyle4] (76.355pt,443.33pt) -- (76.355pt,463.33pt);
        \draw[pstyle5] (72.355pt,453.33pt) -- (76.355pt,463.33pt) -- (80.355pt,453.33pt) -- (76.355pt,457.33pt) -- cycle;
        \draw[pstyle4] (207.475pt,443.33pt) -- (207.475pt,463.33pt);
        \draw[pstyle5] (203.475pt,453.33pt) -- (207.475pt,463.33pt) -- (211.475pt,453.33pt) -- (207.475pt,457.33pt) -- cycle;
        \draw[pstyle4] (343.005pt,443.33pt) -- (343.005pt,463.33pt);
        \draw[pstyle5] (339.005pt,453.33pt) -- (343.005pt,463.33pt) -- (347.005pt,453.33pt) -- (343.005pt,457.33pt) -- cycle;
        \draw[pstyle4] (76.355pt,503.33pt) -- (76.355pt,535.33pt);
        \draw[pstyle4] (76.355pt,535.33pt) -- (199.885pt,535.33pt);
        \draw[pstyle5] (189.885pt,531.33pt) -- (199.885pt,535.33pt) -- (189.885pt,539.33pt) -- (193.885pt,535.33pt) -- cycle;
        \draw[pstyle4] (207.475pt,503.33pt) -- (207.475pt,523.33pt);
        \draw[pstyle4] (207.475pt,523.33pt) -- (211.885pt,523.33pt);
        \draw[pstyle5] (207.885pt,513.33pt) -- (211.885pt,523.33pt) -- (215.885pt,513.33pt) -- (211.885pt,517.33pt) -- cycle;
        \draw[pstyle4] (343.005pt,503.33pt) -- (343.005pt,535.33pt);
        \draw[pstyle4] (343.005pt,535.33pt) -- (223.885pt,535.33pt);
        \draw[pstyle5] (233.885pt,531.33pt) -- (223.885pt,535.33pt) -- (233.885pt,539.33pt) -- (229.885pt,535.33pt) -- cycle;
        \draw[pstyle4] (211.885pt,417.33pt) -- (211.885pt,437.33pt);
        \draw[pstyle5] (207.885pt,427.33pt) -- (211.885pt,437.33pt) -- (215.885pt,427.33pt) -- (211.885pt,431.33pt) -- cycle;
        \draw[pstyle4] (211.885pt,547.33pt) -- (211.885pt,567.33pt);
        \draw[pstyle5] (207.885pt,557.33pt) -- (211.885pt,567.33pt) -- (215.885pt,557.33pt) -- (211.885pt,561.33pt) -- cycle;
        \draw[pstyle4] (211.885pt,607.4399pt) -- (211.885pt,627.4399pt);
        \draw[pstyle5] (207.885pt,617.4399pt) -- (211.885pt,627.4399pt) -- (215.885pt,617.4399pt) -- (211.885pt,621.4399pt) -- cycle;
        \end{tikzpicture}
    }
    \caption{Process of the methods used within this study.}
    \label{fig:methods}
\end{figure}
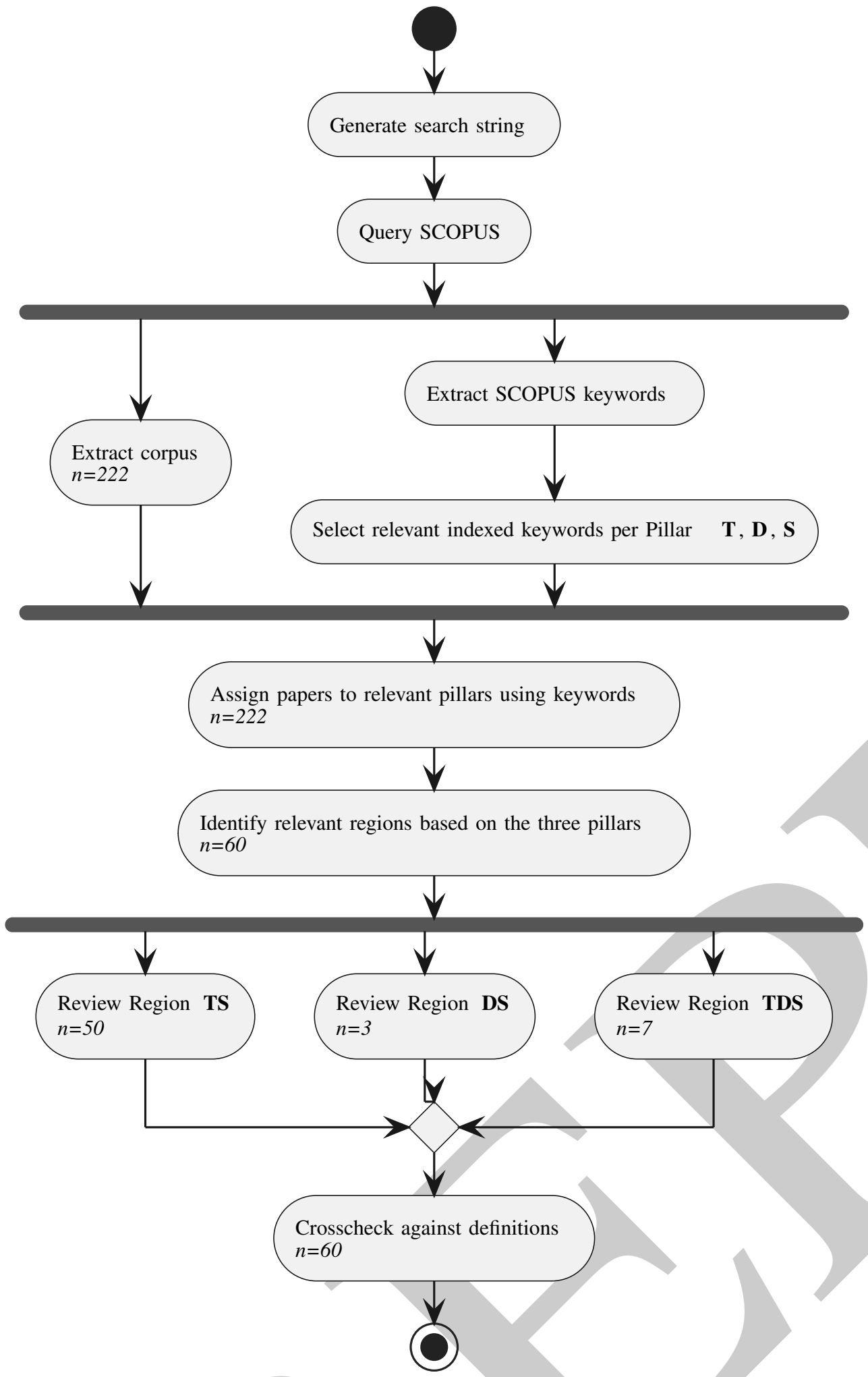

\Cref{sec:methods-rationale} provides a rationale around the choice of our methodological approach, and addresses assumptions and limitations. \Cref{sec:methods-search-string} describes the design of the search string. \Cref{sec:methods-keywords} describes the process of using Scopus indexed keywords to filter the corpus obtained from executing the search string. \Cref{sec:methods-categorising} categorises the literature from the corpus into pillar-aligned \textit{regions}. \Cref{sec:methods-definitions} provides the definitions that are used to cross-check literature against. \Cref{fig:methods} provides a supplementary illustration of the methodological process. This methodology -- including filtering, analysis, and production of figures -- is entirely reproducible; please refer to the supplementary repository to view the code~\cite{preeceSurveypaperhelpers2026}.

\subsection{Rationale, Assumptions, and Limitations for Scopus and Indexed Keyword-Based Filtering} \label{sec:methods-rationale}
Scoping review search methods are commonly limited to database searching and the number of sources may be restricted to those key to the topic~\cite{boothSystematicApproachesSuccessful2022}. We selected Scopus as the sole bibliographic source for two main reasons. First, it provides broad coverage of journal articles, conference proceedings, and book chapters across the three thematic pillars. Second, the methodology adopted in this study relies on the indexed keywords that are computed and provided as metadata by Scopus\footnote{These indexed keywords are assigned by Scopus rather than supplied by the original authors.} to categorise the studies in the corpus against the thematic pillars.

Using Scopus indexed keywords provides a consistent classification mechanism that reduces the degree of subjective interpretation required when assigning studies to the thematic pillars. In contrast, manual classification may introduce greater variability between reviewers, thus reducing reproducibility. While indexed keywords do not fully capture the technical contributions or research objectives of individual studies, deeper analysis is provided after categorisation as part of \Cref{sec:results}.

The use of a single bibliographic database introduces limitations. Studies indexed exclusively in alternative databases, or those not indexed in Scopus, will not have been captured. However, we accepted this trade-off in order to maintain a consistent classification process: incorporating multiple databases would introduce heterogeneity in the keywords, reducing the consistency.

A further limitation arises from the availability of indexed keywords, as noted in \Cref{sec:methods-keywords}. Of the \num{222} retrieved studies, \num{47} did not contain Scopus indexed keywords and were therefore classified using information contained within their abstracts. This approach ensured that potentially relevant studies were not excluded from the review, albeit at the expense of a richer list of keywords defined by Scopus.

Consequently, the study should be interpreted as a Scopus-based scoping review intended to map the structure and characteristics of the literature, rather than provide an exhaustive catalogue of all relevant publications.

\subsection{The Scopus Query Search String} \label{sec:methods-search-string}

\Cref{lst:search-string} shows the search string designed around the three thematic pillars of our study.
\begin{lstlisting}[float=!ht, style=scopus, caption={Scopus database query.}, label={lst:search-string}]
TITLE-ABS-KEY ( 
    ( 
            ontolog* 
        OR  "knowledge graph" 
        OR  semantic* 
    ) 
    AND ( transport* ) 
    AND ( 
            decarbonis* 
        OR  sustain* 
        OR  "net zero" 
        OR  carbon 
    ) 
    AND NOT gene* 
) 
AND PUBYEAR > 2007 
AND PUBYEAR < 2026 
AND ( 
        LIMIT-TO ( DOCTYPE , "ar" ) 
    OR  LIMIT-TO ( DOCTYPE , "cp" ) 
    OR  LIMIT-TO ( DOCTYPE , "ch" ) 
    OR  LIMIT-TO ( DOCTYPE , "re" ) 
) 
AND ( 
        LIMIT-TO ( LANGUAGE , "English" )
)
\end{lstlisting}
The string was left broad to capture a wide range of literature prior to additional filtering, using wildcards where appropriate. 
For decarbonisation\footnote{We cross-checked the American English of the spelling "decarbonization", which returned a corpus of one fewer paper. As such, we retained the British spelling.}, we captured a wider range by also including \scopusterm{sustain*} and \scopusterm{carbon}.
For semantic technologies, \scopusterm{ontolog*}, \scopusterm{knowledge graph}, and \scopusterm{semantic*} were the key terms. The wildcard \texttt{NOT gene*} was included in order to remove literature associated with the field of genetics, given the widespread use of ontologies in biomedical research. 

\begin{figure*}[ht!]
    \centering
    \includegraphics[width=1\linewidth]{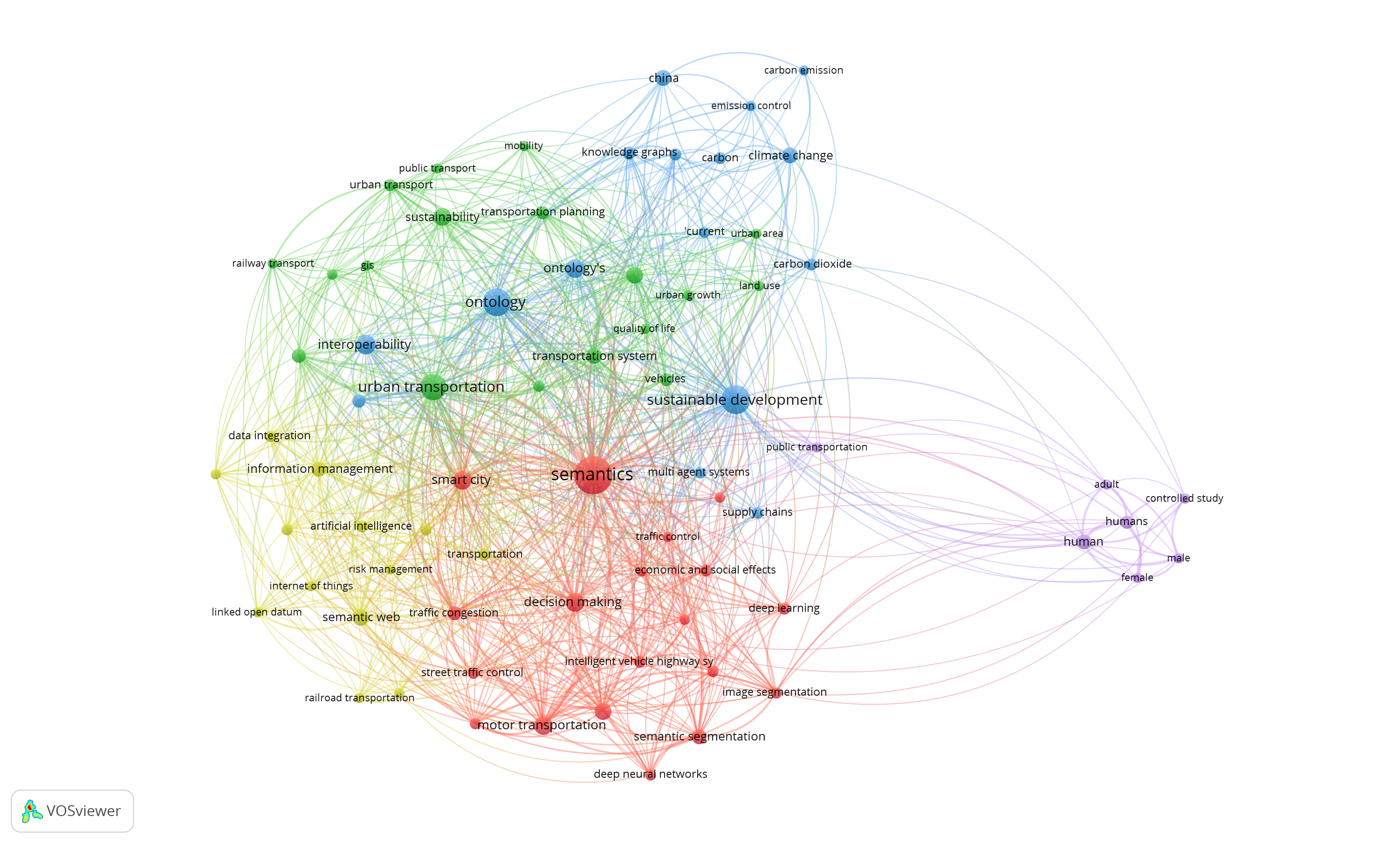}
    \caption{Visualisation of Scopus indexed keyword co-occurrence, created in VOSviewer. More frequent keywords have larger circles and labels, lines represent co-occurrences of keywords, and the distance between keywords approximates their relatedness~\cite{steeleDeterminingFutureHigh2024}
    .}
    \label{fig:keyword_co_occurence}
\end{figure*}

We filtered results to studies published after 2007 and before 2026. Although \gls{owl} became a \gls{w3c} Recommendation in 2004~\cite{bechhoferOWLWebOntology2004}, the selected starting point more closely reflects the period during which \gls{owl}[\space2]~\cite{grauOWL2Next2008} and related ontology engineering approaches became more widely adopted across academic and industrial applications. As this review focuses on the application of semantic technologies within transport and decarbonisation research, we considered this period to better reflect the broader uptake of these technologies beyond the core Semantic Web community.

We limited document types to articles, conference papers, book chapters, and review papers. Furthermore, we limited results to those written in English. 

The search yielded \num{222} studies\footnote{This value was correct as of \DTMdate{2026-03-26}, when the search was executed.}. To assess the sensitivity of the query, we removed each thematic pillar in turn: removing transport-related terms yielded \num{7009} studies; removing decarbonisation terms yielded \num{3979} studies; and removing semantic technology terms yielded \num{179923} studies. This highlights the comparatively narrow overlap between the three thematic pillars. When the exclusion criterion \scopusterm{NOT gene*} is omitted, the search returned \num{1035} studies, indicating significant overlap with biomedical literature. These larger result sets confirm that the final corpus of \num{222} studies represents an appropriately scoped intersection of the three thematic pillars. The earliest relevant studies appear in 2010, supporting the decision to restrict the search window to publications from 2007 onwards.

\subsection{Corpus Filtering on Scopus Indexed Keywords} \label{sec:methods-keywords}
To analyse the literature, we group the studies against our three thematic pillars by their Scopus indexed keywords. Scopus indexed keywords are ``chosen by Scopus and are standardized to vocabularies derived from thesauri that Elsevier owns or licenses'' that, unlike author keywords, ``take into account synonyms, various spellings, and plurals''~\cite{elsevierHowAuthorKeywords}. This approach enables studies to be compared on a consistent thematic basis despite differences in author terminology. Once we identified the corpus of \num{222} studies, we extracted the Scopus indexed keywords for each study, giving \num{2258} unique keywords associated with the corpus.

\Cref{fig:keyword_co_occurence} shows the co-occurrence of the Scopus indexed keywords. It was created in VOSviewer\footnote{\href{https://www.vosviewer.com/}{https://www.vosviewer.com/}} using the full counting method and default analysis parameters, with the exception of the normalisation method. This was changed from association strength to LinLog/modularity to better visualise the most frequent keywords. The relationship is shown between the \num{73} keywords that occur more than five times (note that the keyword `article', which meets this criteria, was removed). 

Semantics, urban transportation, sustainable development, and ontology are the most prominent keywords, reflecting balance of the three thematic pillars. They represent the largest three of the five assigned clusters. Clusters, represented by different colours, are automatically created using the \gls{vos} clustering technique. The network and map \texttt{.txt} files are included in the repository~\cite{preeceSurveypaperhelpers2026}, supporting exploration in VOSviewer.

The visualisation shows some overlap between the broader semantics cluster (Cluster 1, red), urban transportation cluster (Cluster 2, green), and ontology and sustainable development cluster (Cluster 3, blue), indicating that a number of studies lie at the intersection of these themes. A fourth cluster concerned with more specific semantic technology techniques (Cluster 4, yellow) also exhibits connections to Clusters 1--3. The fifth cluster, focused on human studies (Cluster 5, purple), is comparatively isolated, as indicated by its greater distance from the remaining clusters.

For each thematic pillar, we determined a subset of keywords from the full list of \num{2258} indexed keywords: \Cref{lst:keywords-semantic,lst:keywords-transportation,lst:keywords-decarbonisation} (in \Cref{sec:filtered-keywords}) show these choices in detail. To assist in selecting pertinent keywords from the full list, the complete \gls{csv} export of indexed keywords was uploaded to Microsoft Copilot Enterprise. Copilot was then prompted to compile candidate keyword lists for each of the three thematic pillars. These candidate lists were reviewed manually by the reviewing authors, who refined, amended, and supplemented the suggestions before agreeing on the final keyword subsets. The keywords presented in \Cref{sec:filtered-keywords} represent the final selections used in this study. For transparency and reproducibility, the original Copilot-generated candidate lists are available in the accompanying GitHub repository~\cite{preeceSurveypaperhelpers2026}.

The motivation to take this approach was to allow us to target literature pertinent to the three thematic pillars as decided by the keyword indexing engine of Scopus, while filtering out those irrelevant to our study. Furthermore, it minimised the impact of reader bias at the categorisation stage by allowing an indexing engine to ascertain relevance. Of the \num{222} studies, \num{47} contained no Scopus indexed keywords: these are identified in the repository~\cite{preeceSurveypaperhelpers2026}. For these studies only, we cross-checked the abstracts against the pillar keyword subsets: if a match was found, the study was included.

\subsection{Categorising the Literature Into the Thematic Pillars for Review} \label{sec:methods-categorising}
\begin{figure}[ht]
    \centering
    \includegraphics[width=\linewidth,trim={0, 1cm, 0, 1cm}, clip]{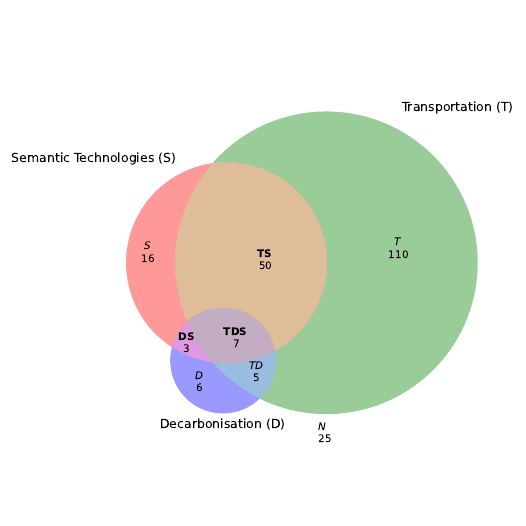}
    \caption{Venn diagram of how the \num{222} studies align with each pillar. Regions selected for review are in bold, and italics otherwise.}
    \label{fig:venn3}
\end{figure}

\Cref{fig:venn3} illustrates the cross-over of the \num{222} studies between the three thematic pillars.
Within this Venn diagram, there are eight regions. \region{N} is the \num{25} studies that have no relevant keywords to the three thematic pillars; upon inspection, these were predominantly biological papers not excluded by the \scopusterm{NOT gene*} filter.

Of the remaining studies, \region{T} has \num{110} studies associated only with transport keywords, \region{S} has \num{16} studies with semantic technology keywords, and \region{D} has \num{6} studies only with decarbonisation keywords. Pairwise overlaps were more limited: \region{TS} has \num{50} studies linking transport and semantic technology keywords; \region{TD} has \num{5} studies linking transport and decarbonisation keywords; and \region{DS} has \num{3} studies linking semantic technology and decarbonisation keyword. \region{TDS} -- studies that have keywords associated with all three thematic pillars -- has just \num{7} studies. This assignment to regions correlates with the keyword co-occurrence. 

\begin{figure}[ht]
    \centering
    \includegraphics[width=1\linewidth]{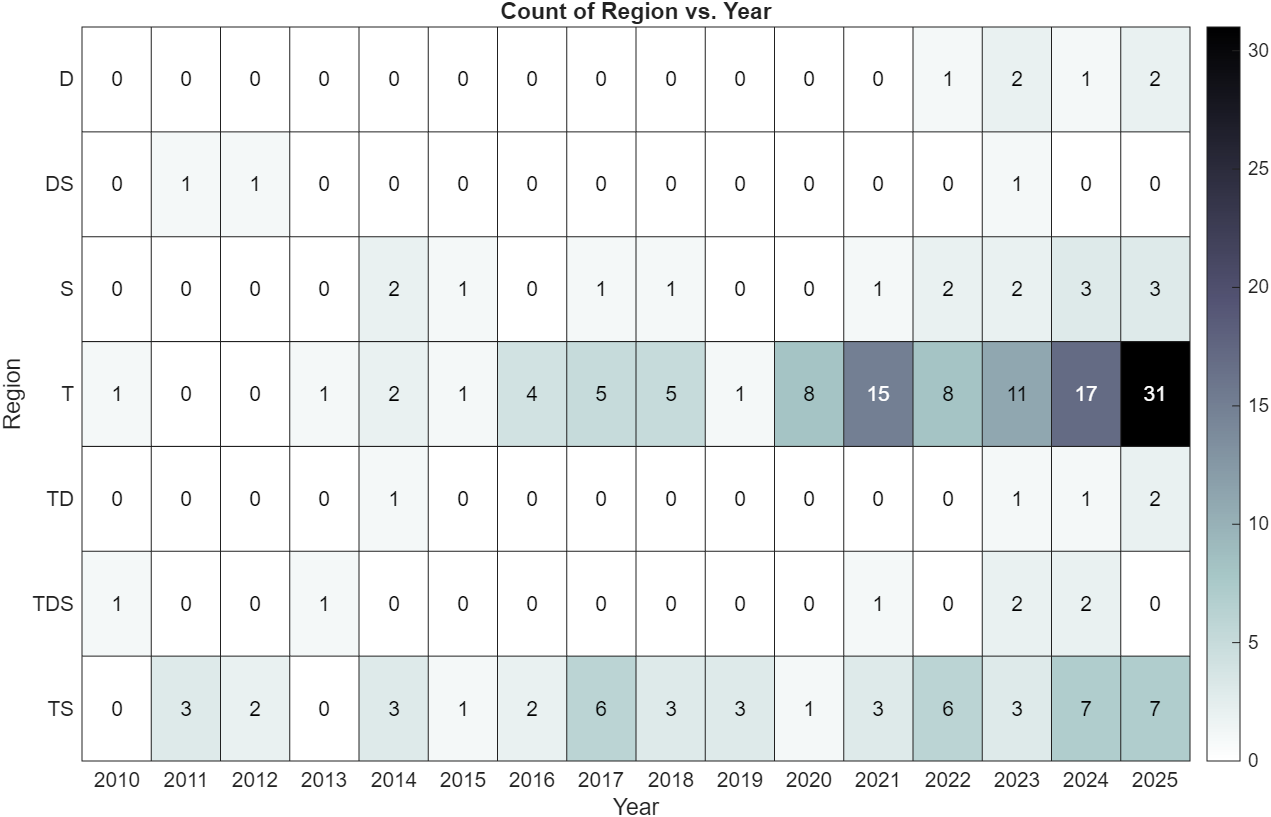}
    \caption{Number of studies published each year within all regions bar \region{N}.}
    \label{fig:heatmap}
\end{figure}

\Cref{fig:heatmap} shows how many studies were published each year within each region. There has been a sharp increase in transport related publications, with over half of the studies assigned to \region{T} published in the last three years. Growth in studies that apply semantic technologies within transport (\region{TS}) has been slower but is also evident. Given the much lower number of studies related to decarbonisation, it is challenging to identify an immediate pattern. However, when \regions{D, DS, TD, and TDS}, are grouped together, it can be seen that two thirds of studies in these regions were published in the last three years, suggesting increased recent interest in decarbonisation research. 

These regions form the basis of our analysis in \Cref{sec:results}. \region{N} (the \num{25} studies beyond the scope of the three thematic pillars) and \regions{T, D, S, and TD} (the \num{137} associated with just a single pillar or not with semantic technologies) were excluded from the final review. \regions{TS, DS, and TDS} -- the remaining \num{60} studies representing the intersections between transport and decarbonisation pillars with semantic technologies -- were retained.

\begin{table*}[ht!]
    \centering
    \caption{Definitions of concepts within -- and related to -- Semantic Web technologies.}
    \begin{tabularx}{\textwidth}{lX}
        \toprule
        Term (general) & Definition \\
        \midrule
        Semantic technology & Semantic technologies comprise a set of methods for representing, integrating, and reasoning over data using formal semantics, enabling machines to interpret relationships and meaning more explicitly. \\ \addlinespace

        Semantic Web technology & Semantic Web technologies are a subset of semantic technologies based on \glsfmtshort{w3c} standards (e.g. \glsfmtshort{rdf}, \glsfmtshort{owl}, \glsfmtshort{sparql}) that enable the representation and querying of structured data across the \glsfmtshort{www}. \\ \addlinespace
        
        Taxonomy & A taxonomy is a hierarchical classification of concepts or entities based on shared characteristics, primarily intended to support organisation and retrieval rather than formal semantic modelling~\cite{brachmanWhatIsaIsnt1983}. \\ \addlinespace

        \midrule
        Term (cross-check criteria) & Definition \\
        \midrule
        Ontology & An ontology is a formal and explicit specification of a domain, including concepts, relationships, and logical axioms that constrain interpretation and support reasoning~\cite{staabHandbookOntologies2013}. \\ \addlinespace

        \Glsfmtfull{kg} & A knowledge graph is a graph-based data structure that represents entities (nodes) and their relationships (edges), often enriched with semantics to enable integration and querying. \\ \addlinespace
        
        \Glsfmtshort{owl} ontology & An \glsfmtshort{owl} ontology is an ontology expressed using \glsfmtshort{owl}~\cite{bechhoferOWLWebOntology2004,grauOWL2Next2008}, a formal language grounded in \glsfmtlong{dl} that enables well-defined semantics and automated reasoning. \\ \addlinespace 
        
        \glsfmtshort{rdf} \glsfmtlong{kg} & 
        An \glsfmtshort{rdf} knowledge graph is a knowledge graph represented using the \glsfmtshort{rdf} data model, in which information is expressed as triples (subject--predicate--object). \\ \addlinespace
        
        Artefact & An artefact is a publicly accessible file, repository, or document that shares the semantic technology described in a study in accordance with \glsfmtshort{fair} principles. \\
        
        \bottomrule 
    \end{tabularx}
    \label{tab:definitions}
\end{table*}

It is important to note that within these regions, the studies are still contextualised by the initial search string. For example, \region{TS} has \num{50} studies that use semantic technologies for transport without addressing \textit{decarbonisation}, but are still among a corpus that address a form of \textit{sustainability}. 

\subsection{The Definitions of the Semantic Technologies Terminology} \label{sec:methods-definitions}
To support interpretation and provide a common baseline, \Cref{tab:definitions} provides definitions that we use in the remainder of the study. This includes definitions of wider semantic technology concepts (such as ontology and knowledge graph) and specific \gls{w3c} Semantic Web terms (such as \gls{owl} ontology and \gls{rdf} knowledge graph). 

The separation between wider semantic technology terms and specific Semantic Web technology terms is a key consideration when reviewing the literature. For example, consider \textit{ontology} and \textit{\gls{owl} ontology}: where \textit{ontology} may be used to address philosophical ontologies, general formalisations of concepts, or \gls{owl}[-based] ontologies, \textit{\gls{owl} ontology} only addresses ontologies that comply with the standards of the \gls{w3c}. It is crucial to differentiate these, as the adherence to widely-adopted standards is a key way to address semantic heterogeneity. This forms the basis of our analysis, as each of the \num{60} studies is cross-checked against whether it is \gls{w3c} compliant or otherwise, and whether a publicly accessible artefact is provided. We emphasise that these results are based on what the authors presented in the original work.

\section{Results} \label{sec:results}

This section analyses the \num{60} studies retained from \Cref{sec:methods}, structured according to the regions in \Cref{fig:venn3}: \Cref{sec:results-ts} covers \region{TS} (50 papers); \Cref{sec:results-ds} covers \region{DS} (\num{3} papers); and \Cref{sec:results-tds} covers \region{TDS} (\num{7} papers).

The analysis is targeted towards answering \Cref{rq1,rq2,rq3}. Within each section we include a table indicating how the relevant studies align with the cross-check criteria related to semantic technologies (\Cref{rq1}), Semantic Web technologies (\Cref{rq2}), and which provide publicly available artefacts (\Cref{rq3}).

\subsection{\region{TS}: Transport and Semantic Technologies} \label{sec:results-ts}

\region{TS} contains \num{50} of the \num{60} selected studies. For clearer analysis, we split this section further into subthemes based on Scopus indexed keywords in \Cref{lst:keywords-semantic-web,lst:keywords-ontology,lst:keywords-kg,lst:keywords-lod}: \Cref{sec:results-ts-semantic-web} addresses studies associated with \textit{Semantic Web} Scopus keywords in \Cref{lst:keywords-semantic-web} (\num{15} studies); \Cref{sec:results-ts-ontology} addresses studies associated with \textit{ontology} Scopus keywords in \Cref{lst:keywords-ontology}  (\num{29} studies); and \Cref{sec:results-ts-kg} addresses studies associated with \textit{knowledge graph} Scopus keywords in \Cref{lst:keywords-kg} (\num{10} studies). Note that there are four studies associated with multiple subthemes based on their keywords; we review them only in the first subtheme they appear in, but indicate each subtheme they appear in within the associated tables.

\subsubsection{Semantic Web Scopus Keywords} \label{sec:results-ts-semantic-web}

Several studies in the transport domain adopt an integrated use of established \gls{w3c} Semantic Web technologies and principles, combining \gls{rdf}[-based] storage, \gls{owl} ontologies, and \gls{lod} principles. Some studies employ multiple technologies from the Semantic Web stack, whereas some focus on an individual technology. Many of these studies provide artefacts to compliment their work. This suggests that a subset of literature adheres to the existing standards endorsed by \pgls{w3c}. This is significant, as it means research outputs are aligned with the \gls{fair} principles, and can be adapted and expanded more easily. 
\Cref{tab:results-ts-semantic-web} indicates how these studies align with the definitions presented in \Cref{tab:definitions}.
\begin{table}[ht]
    \rowcolors{2}{gray!15}{white}
    \centering
    \caption{The studies within \region{TS} (under the additional ``Semantic Web'' keywords provided in \Cref{lst:keywords-semantic-web}) cross-checked against \Cref{tab:definitions}. These papers were reviewed by Haoruo Zhao and Duncan Guthrie.}
    \label{tab:results-ts-semantic-web}
    \resizebox{\columnwidth}{!}{%
    \begin{tabular}{lccccc}
        \toprule
        Study & \rotatebox[origin=l]{90}{Ontology} & \rotatebox[origin=l]{90}{\glsentryshort{kg}} & \rotatebox[origin=l]{90}{\glsentryshort{owl} Ontology} & \rotatebox[origin=l]{90}{\glsentryshort{rdf} \glsentryshort{kg}} & \rotatebox[origin=l]{90}{Artefact} \\
        
        \midrule
        \multicolumn{6}{c}{\textit{Combining multiple Semantic Web technologies}}\\
        \midrule

        \textcite{consoliEmergencyVehicleRouting2014a} & \checkmark & \checkmark & \checkmark & \checkmark & \checkmark \\
        \textcite{consoliSettingCourseEmergency2014a} & \checkmark & \checkmark & \checkmark & \checkmark & \checkmark \\
        \textcite{colacinoManagingRoadSafety2017a} & \checkmark & \checkmark & \checkmark & \checkmark & \checkmark \\
        \textcite{palSupplyChainInformation2023} & \checkmark & \checkmark & \checkmark & \checkmark &  \\

        \midrule
        \multicolumn{6}{c}{\textit{Focused on individual technologies}}\\
        \midrule
         
        \textcite{golinskaAgentbasedSystemPlanning2011a} & \checkmark & \checkmark & \checkmark & \checkmark &  \\
        \textcite{caceresKnowledgeModelingSustainable2014a} &  & \checkmark &  & \checkmark &  \\
        \textcite{buyleSustainableMethodPublishing2021a} &  & \checkmark &  & \checkmark &  \\
        \textcite{liuBridgingDataSilos2025a} & \checkmark & \checkmark & \checkmark & \checkmark & \checkmark \\

        \midrule
        \multicolumn{6}{c}{\textit{Using \glsxtrshort{owl}}}\\
        \midrule
        \textcite{lecueApplyingSemanticWeb2012a} & \checkmark & \checkmark & \checkmark & \checkmark & \checkmark \\
        \textcite{bouhanaOntologybasedCBRApproach2015a} & \checkmark &  & \checkmark &  &  \\

        \midrule
        \multicolumn{6}{c}{\textit{Exploratory and conceptual work}}\\
        \midrule
        \textcite{gogosIT2RailInformationTechnologies2016a} & \checkmark & \checkmark & \checkmark & \checkmark & \checkmark \\
        \textcite{ngMasterDataManagement2017a} & \checkmark & \checkmark & \checkmark & \checkmark &  \\  

        \midrule
        \multicolumn{6}{c}{\textit{Beyond scope}} \\
        \midrule
        \textcite{bouchereauSemanticAnnotationResources2017a} &  &  &  &  &  \\
        \textcite{masekResearchPassengersDemand2018a} &  &  &  &  &  \\
        \textcite{gaudet-blavignacNationalSemanticdrivenThreepillar2021a} &  &  &  &  &  \\

        \bottomrule
    \end{tabular}
    }
\end{table}

\paragraph{Studies Combining Multiple Semantic Web Technologies} \label{sec:results-ts-semantic-web-multiple}
Four studies combine multiple Semantic Web technologies to address specific challenges.
\textcite{consoliEmergencyVehicleRouting2014a,consoliSettingCourseEmergency2014a} transform heterogeneous municipal \gls{gis} data into interoperable knowledge resources. In particular, \gls{rdf} and an \gls{owl} ontology are used to formalise entities, properties, and spatial relations extracted from the territorial information system of Catania, while \gls{lod} principles and existing vocabularies support interoperability across datasets. Shape-based geo-data are converted into \gls{rdf} triples through \gls{sparql}-based transformation scripts, and the resulting knowledge graph is exposed through \gls{sparql} endpoints to enable integration and querying of urban mobility and emergency-routing data. 

Similarly, \textcite{colacinoManagingRoadSafety2017a} use \gls{owl} and \gls{rdf} to transform Italian road-accident statistics into \gls{lod} integrated with external geographic resources and OpenStreetMap, supporting semantic integration and map-based analysis of accidents, fatalities, and injuries. Furthermore, \textcite{palSupplyChainInformation2023} propose a semantic data-integration framework for supply-chain information systems aimed at improving sustainability and interoperability across heterogeneous business data sources through a mediating ontology layer. \gls{rdf} graphs are used to represent integrated knowledge, while \gls{owl} supports richer ontology modelling beyond the expressiveness of \gls{rdf}/\gls{rdfs}, particularly for more complex semantic relationships. \gls{sparql} is further employed to query the integrated semantic layer and support query rewriting across mapped relational and \gls{xml} data sources.

\paragraph{Studies Focused on Individual Technologies}
While the studies above combine multiple Semantic Web technologies, four other works focus more specifically on individual technologies. \gls{rdf}, in particular, is widely used to support interoperability and semantic integration across heterogeneous transport data sources. Several transport studies employ \gls{rdf}[-based] approaches to represent, integrate, and exchange transport-related information. \textcite{golinskaAgentbasedSystemPlanning2011a} use \gls{rdf} to define a schema for describing an agent-based model of a multimodal transport system, with particular emphasis on cooperation between regional transport sectors. \textcite{caceresKnowledgeModelingSustainable2014a} describe the use of the \gls{rdf}-based Transmodel and \gls{ifopt} models to support reasoning about sustainable public transport. The work emphasises interoperability across meta-models, showing how limitations of Transmodel in representing bus transfers can be addressed through \gls{ifopt}['s] more general conceptualisation of places and points. It further highlights the use of semantic modelling to uncover additional transport phenomena through the conceptual interpretation of real-world data. 

In addition, \textcite{buyleSustainableMethodPublishing2021a} apply \gls{rdf} and \gls{lod} principles to publish interoperable open data on air quality and railway infrastructure. Similarly, \textcite{liuBridgingDataSilos2025a} develops the \gls{evca} ontology to integrate heterogeneous data on \gls{ev} charging infrastructure, neighbourhood characteristics, and human mobility, and implements this semantic framework by converting multi-source datasets into \gls{rdf} triples for cross-domain querying and interoperability. It demonstrates, through a case study in Eindhoven, that semantic modelling can support integrated analyses of charging accessibility, operator distribution, and mobility-informed infrastructure planning, although the validation remains based on a relatively limited dataset.

\paragraph{Studies Using \glsfmtlong{owl}}
Within Semantic Web technologies, \gls{owl} ontologies provide a formal layer for knowledge modelling and automated reasoning built upon \gls{rdf}[-based] data representation, and have consequently attracted considerable attention in transport applications for representing complex relationships and constraints within transport knowledge.
\textcite{lecueApplyingSemanticWeb2012a} used Dublin, Ireland as an application scenario for road congestion diagnosis and show that a \gls{owl-el} -- a lightweight \gls{owl} profile based on the \glspl{dl} $\mathcal{EL}$++~\cite{baaderPushingEnvelope2005} -- provides a feasible formal basis for representing the relevant traffic knowledge. Their framework combines an \gls{owl-el} ontology with stream and historical data in order to explain persistent congestion, rather than merely detect it. The work illustrates how the restriction to \gls{owl-el} can support efficient reasoning while retaining clear practical relevance. 

Furthermore, \textcite{bouhanaOntologybasedCBRApproach2015a} propose an ontology-supported case-based reasoning framework for personalised itinerary search in urban freight transport. The approach combines an \gls{owl} ontology with \gls{swrl} rules to support personalised case representation and inference over stakeholder preferences, transport objectives, and itinerary constraints. \gls{swrl} rules are further employed to capture user-specific requirements and guide semantic case retrieval, illustrating how rule-based reasoning can complement ontology-based transport knowledge representation. 

\paragraph{Exploratory and Conceptual Work}
The potential of Semantic Web technologies for transport systems has also attracted attention at a more conceptual and exploratory level. Three studies discuss the prospective role of Semantic Web technologies in supporting future interoperability, integration, and intelligent transport services, often from the perspective of architectural proposals, design visions, or preliminary user-oriented investigations rather than fully implemented systems. For example, \textcite{ngMasterDataManagement2017a} present a master data-management framework for smart-city infrastructure data, where Semantic Web technologies such as \gls{rdf} and \gls{owl} are discussed primarily as supporting technologies and possible implementation mechanisms. 

\textcite{gogosIT2RailInformationTechnologies2016a} further present an architectural vision for interoperable multimodal travel services in which Semantic Web technologies play a central enabling role. In particular, the work outlines an ontology-based framework together with a shared ontology repository to support semantic interoperability across distributed travel-related services and data sources.

\paragraph{Beyond Scope}
Two studies identified are not directly situated within the transport domain, and are therefore beyond the scope of our transport-focused survey, although they contribute to the broader Semantic Web literature. \textcite{bouchereauSemanticAnnotationResources2017a} investigate the use of Semantic Web technologies in an e-learning context, while \textcite{gaudet-blavignacNationalSemanticdrivenThreepillar2021a} focus on semantic and ontology-based approaches in the biomedical domain.

\textcite{masekResearchPassengersDemand2018a}, associated with the \gls{gof4r} project, refer to semantic technologies in the context of interoperability, although the paper itself mainly reports ethnographic workshops investigating passenger requirements for a travel companion system, and does not present any direct discussion of ontologies or knowledge graphs.

\subsubsection{Ontology Scopus Keywords} \label{sec:results-ts-ontology}
Many studies use ontologies to support interoperable knowledge representation, integration, and reuse across heterogeneous systems and datasets. In particular, \gls{owl} ontologies often exist alongside \gls{rdf} knowledge graphs and are used as part of larger decision support systems. In this context, ontologies can control domain vocabularies and enable the integration of data from heterogeneous sources~\cite{carbonaroFrameworkDataConcept2019a, badiiAnalysisAssessmentKnowledge2017a}. Other forms of data-driven ontology development that do not employ \gls{owl} or related \gls{w3c} standards also appear in this region. Although such approaches are relevant to ontology engineering, the lack of alignment with widely adopted Semantic Web standards may limit interoperability and reuse. Additionally, this region contains philosophical and socio-theoretical ontologies. While these contributions may be of interest to the broader ontology community, they fall outside the scope of the present analysis. \Cref{tab:results-ts-ontology} indicates which of these studies aligns with the definitions presented in \Cref{tab:definitions}.
\begin{table}[ht!]
    \rowcolors{2}{gray!15}{white}
    \centering
    \caption{The studies within \region{TS} (under the additional ``ontology'' keywords provided in \Cref{lst:keywords-ontology}) cross-checked against \Cref{tab:definitions}. Studies already reviewed as part of \Cref{sec:results-ts-semantic-web} are italicised. These papers were reviewed by Joseph D. Preece and Haoruo Zhao.}
    \label{tab:results-ts-ontology}
    \resizebox{\columnwidth}{!}{%
    \begin{tabular}{lccccc}
        \toprule
        Study & \rotatebox[origin=l]{90}{Ontology} & \rotatebox[origin=l]{90}{\glsentryshort{kg}} & \rotatebox[origin=l]{90}{\glsentryshort{owl} Ontology} & \rotatebox[origin=l]{90}{\glsentryshort{rdf} \glsentryshort{kg}} & \rotatebox[origin=l]{90}{Artefact} \\
        \midrule

        \textcite{carbonaroFrameworkDataConcept2019a} & \checkmark &  & \checkmark &  &  \\

        \midrule
        \multicolumn{6}{c}{\textit{Ontologies for decision support systems}}\\
        \midrule

        \textcite{osmanKnowledgeenabledDecisionSupport2011a} & \checkmark & \checkmark & \checkmark & \checkmark &  \\
        \textcite{badiiAnalysisAssessmentKnowledge2017a} & \checkmark & \checkmark & \checkmark & \checkmark & \checkmark \\
        \textcite{memonSemanticTransportationPlanning2017a} & \checkmark & \checkmark & \checkmark & \checkmark &  \\
        \textcite{moskolaiOntologyBasedApproach2019} & \checkmark &  & \checkmark &  &  \\
        \textcite{antoniouEnablingUsePlanning2019a} & \checkmark & \checkmark & \checkmark & \checkmark &  \\
        \textcite{preeceLeveragingOntochainsDistributed2024b} & \checkmark & \checkmark & \checkmark &  &  \\
        \textcite{martinezVRUSafetyRoad2025a} & \checkmark &  & \checkmark &  &  \\

        \midrule
        \multicolumn{6}{c}{\textit{Ontologies and stand-alone artefacts}}\\
        \midrule

        \textcite{anandGenCLOnOntologyCity2012a} & \checkmark & \checkmark & \checkmark & \checkmark &  \\
        \textcite{caoStudyRiskRelevance2020a} & \checkmark & \checkmark & \checkmark & \checkmark &  \\
        \textcite{saafiOntologicalModelEnhance2024a} & \checkmark &  & \checkmark &  &  \\
        \textcite{bhuyanOntologyDevelopmentSustainable2024a} & \checkmark &  &  &  &  \\
        \textcite{asgariOntologyResilienceSustainability2025a} & \checkmark & \checkmark & \checkmark & \checkmark &  \\

        \midrule
        \multicolumn{6}{c}{\textit{Ontologies beyond \glsxtrshort{w3c} standards}}\\
        \midrule

        \textcite{smirnovDevelopmentBrokerLogic2011a} & \checkmark &  &  &  &  \\
        \textcite{hoDesignMultiagentEmergency2016} & \checkmark & \checkmark & \checkmark & \checkmark &  \\
        \textcite{giretOntologySustainableIntelligent2018} & \checkmark &  &  &  &  \\
        \textcite{sinhaFrameworkEnsuringZero2022a} & \checkmark &  &  &  &  \\
        \textcite{benesKnowledgeGraphsTransport2023a} & \checkmark & \checkmark &  &  &  \\
        \textcite{curranFoundationsTransdisciplinaryEngineering2024a} & \checkmark & \checkmark &  &  &  \\
        \textcite{tavakkoliUnveilingSynergiesIntegrating2024a} & \checkmark &  &  &  &  \\
        \textcite{feketeOntologyguidedCausalDiscovery2025a} & \checkmark & \checkmark &  &  & \checkmark \\

        \midrule
        \multicolumn{6}{c}{\textit{Review papers}}\\
        \midrule

        \textcite{xiaStudyCityDigital2022b} & \checkmark & \checkmark & \checkmark & \checkmark &  \\

        \midrule
        \multicolumn{6}{c}{\textit{Beyond scope}}\\
        \midrule

        \textcite{zhangUseComparativeQuantitative2017a} &  &  &  &  &  \\
        \textcite{shellerTheorizingMobilityJustice2018a} &  &  &  &  &  \\
        \textcite{kimbrellElectricCarsharingSustainable2021a} &  &  &  &  &  \\
        \textcite{denissenRequirementsCorrosionInhibitor2022} &  &  &  &  &  \\
        \textcite{parkChangingIslandSociety2022a} &  &  &  &  &  \\

        \midrule
        \multicolumn{6}{c}{\textit{Reviewed as part of \Cref{sec:results-ts-semantic-web}}}\\
        \midrule
        
        \textit{\textcite{palSupplyChainInformation2023}} & \checkmark & \checkmark & \checkmark & \checkmark &  \\
        \textit{\textcite{liuBridgingDataSilos2025a}} & \checkmark & \checkmark & \checkmark & \checkmark & \checkmark \\
        \bottomrule
    \end{tabular}
    }
\end{table}

\paragraph{Ontologies for Decision Support Systems}
\textcite{osmanKnowledgeenabledDecisionSupport2011a} present a system to support the selection of the most suitable route for buried urban utilities such as gasworks. Their ontology defines the relevant types of entities in the surrounding environment -- including traffic conditions -- to support the decision‑making process. \textcite{memonSemanticTransportationPlanning2017a} address transport planning for food supply chains in ``geographically difficult'' zones. They develop \gls{ipoves}, a multi‑agent system that enables cooperation between carriers and shippers. As part of this broader system, they construct a hierarchical ontology for food products, a transporter system ontology, and a global mid‑level ontology. \gls{swrl} is used to help constrain data in the ontology‑based approach.

In addition, \textcite{badiiAnalysisAssessmentKnowledge2017a} develop a ``smart-city architecture'' to enable better integration of heterogenous \glspl{api}, grounded around the pre-existing Km4City ontology. \textcite{moskolaiOntologyBasedApproach2019} apply \gls{owl} ontologies for complexity management in the design of sustainable urban mobility systems. Similarly, \textcite{antoniouEnablingUsePlanning2019a} employ a planning agent for urban traffic management, with an \gls{owl} ontology playing a central role. specific

Furthermore, \textcite{preeceLeveragingOntochainsDistributed2024b} presents \gls{stub} 2.0, a distributed public transit ticketing framework that integrates an \gls{owl}[-based] transport network ontology with blockchain to support secure, transparent, and interoperable ticket issuance and validation. By representing transport services, stops, and planned stops in an ontology and linking this semantic model to on-chain smart contracts through ontochains and Merkle-proof-based validation, the system improves ticket verification across multi-operator transport networks. \textcite{martinezVRUSafetyRoad2025a} proposes an ontology-driven digital twin for road intersections, using an \gls{owl} ontology to model both the static infrastructure and dynamic traffic elements that affect \gls{vru} safety. It further demonstrates how this semantic model can be instantiated with open-source tools and smart data models to support visualisation, simulation, and risk analysis for safer intelligent transport systems.

\paragraph{Ontologies and Stand-Alone Artefacts}
Other studies produce the ontology as the artefact in focus. \textcite{anandGenCLOnOntologyCity2012a} produce \gls{genclon}, which is designed to formalise concepts around agent-based models so that modellers can standardise their models. \textcite{caoStudyRiskRelevance2020a} introduce an integrated evolutionary model called \gls{srac}, that employs \gls{owl}/\gls{rdf} representations alongside reasoning capabilities. TrafCsOnto is introduced by \textcite{saafiOntologicalModelEnhance2024a} as an \gls{owl} ontology for smart city traffic management, which models key concepts such as incidents, infrastructure, sensors, persons, trajectories, and vehicles. Its main contribution is a richer conceptual structure for interoperability and traffic condition representation, while practical real-time deployment is identified as future work. 

Furthermore, \textcite{bhuyanOntologyDevelopmentSustainable2024a} develops an \gls{owl} ontology for sustainable intelligent transport systems and represents it using \gls{rdf}[-compatible] semantic standards to support interoperability, reasoning, and real-time decision-making. A notable strength of the work is that the ontology is not presented only as a conceptual model, but is linked to \gls{sumo}[-based] traffic simulation and dynamic data integration for validation in practical traffic management scenarios. \textcite{asgariOntologyResilienceSustainability2025a} presents IWT2Onto, an \gls{owl} ontology for inland waterway transport that models network infrastructure, climate-related hazards, and resilience/sustainability \glspl{kpi} in a unified semantic framework. It further shows how the ontology is populated with open and real-time data in a \gls{rdf} knowledge graph to support \gls{kpi} assessment and decision-making, such as monitoring navigability and identifying alternative routes under flood conditions.

\paragraph{Ontologies Beyond \glsfmtlong{w3c} Standards}
Some studies present ontology‑related research but do not do so within the context of \gls{w3c} standards. \textcite{smirnovDevelopmentBrokerLogic2011a}, for example, develop a broker‑logic system for ride‑sharing built on the Smart‑M3 platform. While they claim to use an ontology to describe the major system components, they provide no indication that it is expressed in \gls{rdf} or \gls{owl}, possibly preventing reusability under the \gls{fair} principles.

Although \textcite{curranFoundationsTransdisciplinaryEngineering2024a} introduces a \gls{tko}, the term ontology is used in a conceptual knowledge-structuring sense. It should be regarded as a conceptual transdisciplinary knowledge framework instead of an ontology engineering study.

In addition, \textcite{hoDesignMultiagentEmergency2016} design an ``ontology‑based'' dispatching model to enhance the resilience of Taiwan’s high‑speed rail system; although, the ontology does not appear to be implemented using \gls{owl} or \gls{rdf}. \textcite{giretOntologySustainableIntelligent2018} explicitly propose an ontology for sustainable intelligent transport. The ontology is presented as a high‑level \gls{uml}[‑like] conceptual model rather than a formal \gls{owl}/\gls{rdf} artefact. While the transport relevance of this work is clear, it does not meet the technical criteria required for inclusion here. \textcite{benesKnowledgeGraphsTransport2023a} proposes a conceptual framework for linking transport, emissions, and meteorological conditions in Prague.

\textcite{tavakkoliUnveilingSynergiesIntegrating2024a} presents a conceptual interoperability framework that aligns \gls{ieee} 1451 and \gls{iso} 15926 for smart manufacturing and transport. Although it refers to ontology mapping and semantic reasoning, the ontological component remains largely high-level, with limited detail on the concrete modelling, alignment, and reasoning process. \textcite{sinhaFrameworkEnsuringZero2022a} propose an ontology-informed framework for analysing sustainability and defects in major Indian ports by linking operational performance with undesirable outputs such as emissions and energy use. It introduces a port-sustainability index (PSI) that combines factors such as vessel load, draft constraints, and turnaround time, and uses competency questions to identify defects and derive policy interventions. \textcite{feketeOntologyguidedCausalDiscovery2025a} use a lightweight ontology-guided constraint framework to improve causal discovery and inference for transport-related \ce{CO_2} emissions. It shows that domain-informed constraints can reduce spurious causal links and improve the interpretability of causal graphs, supporting more reliable analysis of how vehicle attributes affect emissions.

\paragraph{Review Papers}
The growing potential of ontology-based approaches in transport and smart-city systems has also motivated review and survey studies. Within our corpus, \textcite{xiaStudyCityDigital2022b} review the integration of \glspl{gis} and \gls{bim} in the context of sustainable smart cities, with particular attention to ontology-based approaches for addressing problems of data modelling, interoperability, and semantic integration between \gls{bim} and \gls{gis} standards. Transport, especially urban rail transit, is discussed as one of the application settings in which such semantic integration is relevant for planning and lifecycle management.

\paragraph{Beyond Scope}
Several studies adopt philosophical or social‑theoretical conceptions of ontology to examine themes such as mobility justice~\cite{shellerTheorizingMobilityJustice2018a}, sustainable mobility transitions within actor–network frameworks~\cite{kimbrellElectricCarsharingSustainable2021a}, and the reshaping of communities through land‑to‑island bridge infrastructures~\cite{parkChangingIslandSociety2022a}. Although their inclusion reflects a broader scholarly interest in the conceptualisation of transport within sustainability discourses, these works are not directly pertinent to the research questions addressed here, as they do not engage with ontologies in a data‑ or standards‑oriented sense. 
In addition, \textcite{zhangUseComparativeQuantitative2017a,denissenRequirementsCorrosionInhibitor2022} are beyond scope as they are not transport-related.

\subsubsection{Knowledge Graph Scopus Keywords} \label{sec:results-ts-kg}
This section reveals a mixed landscape in which formally semantic, standards-based approaches coexist with broader data-driven uses of the term ``knowledge graph''. While some studies build on \gls{rdf} and ontology-based representations compatible with the Semantic Web stack, others use knowledge-graph terminology primarily to describe graph-structured data integration or analytics without the same degree of formal semantic commitment. Overall, the literature suggests growing interest in graph-based transport intelligence, but also highlights terminological variation and uneven alignment with \gls{w3c} Semantic Web standards. \Cref{tab:results-ts-kg} indicates which of these studies aligns with the definitions presented in \Cref{tab:definitions}.
\begin{table}[ht!]
    \rowcolors{2}{gray!15}{white}
    \centering
    \caption{The studies within \region{TS} (under the additional ``knowledge graph'' keywords provided in \Cref{lst:keywords-kg}) cross-checked against \Cref{tab:definitions}. Studies already reviewed as part of \Cref{sec:results-ts-ontology} are italicised. These papers were reviewed by Navya Nayak.}
    \label{tab:results-ts-kg}
    \resizebox{\columnwidth}{!}{%
    \begin{tabular}{lccccc}
        \toprule
        Study & \rotatebox[origin=l]{90}{Ontology} & \rotatebox[origin=l]{90}{\glsentryshort{kg}} & \rotatebox[origin=l]{90}{\glsentryshort{owl} Ontology} & \rotatebox[origin=l]{90}{\glsentryshort{rdf} \glsentryshort{kg}} & \rotatebox[origin=l]{90}{Artefact} \\
        
        \midrule
        \multicolumn{6}{c}{\textit{Using \glsxtrshort{rdf} knowledge graphs}}\\
        \midrule

        \textcite{htunIntegratingKnowledgeGraphs2024a} & \checkmark & \checkmark & \checkmark & \checkmark &  \\
        \textcite{brasoveanuExplainableMobilityPrediction2025} & \checkmark & \checkmark & \checkmark & \checkmark &  \\

        \midrule
        \multicolumn{6}{c}{\textit{Using other forms of knowledge graph}}\\
        \midrule

        \textcite{huIndividualTravelKnowledge2022a} &  & \checkmark &  &  &  \\
        \textcite{ahmedKnowledgeGraphBased2022a} &  & \checkmark &  &  &  \\
        \textcite{youFederatedPlatformEnabling2023a} &  & \checkmark &  &  &  \\
        \textcite{yanResearchSustainableDevelopment2025a} & \checkmark & \checkmark &  &  &  \\

        \midrule
        \multicolumn{6}{c}{\textit{Review papers}}\\
        \midrule

        \textcite{rajanKnowledgeGraphsApplications2024a} &  & \checkmark &  &  &  \\

        \midrule
        \multicolumn{6}{c}{\textit{Beyond scope}}\\
        \midrule
        \textcite{tomassiHealthcareSupplyChain2025a} &  &  &  &  &  \\

        \midrule
        \multicolumn{6}{c}{\textit{Reviewed as part of \Cref{sec:results-ts-ontology}}}\\
        \midrule

        \textit{\textcite{benesKnowledgeGraphsTransport2023a}} & \checkmark & \checkmark &  &  &  \\
        \textit{\textcite{asgariOntologyResilienceSustainability2025a}} & \checkmark & \checkmark & \checkmark & \checkmark &  \\
        \bottomrule
    \end{tabular}
    }
\end{table}

\paragraph{Studies Using \glsfmtshort{rdf} Knowledge Graphs}
Among the reviewed works, only two adopt a fully semantic approach grounded in \gls{w3c} standards. \textcite{htunIntegratingKnowledgeGraphs2024a} provides a survey of knowledge graph applications in autonomous vehicle systems, emphasising the role of ontologies and \gls{rdf}[-based] representations in enabling semantic understanding, perception, and decision-making. Although this study does not present a concrete implementation, it offers a detailed conceptual foundation for semantic knowledge graphs, highlighting the importance of structured representations and reasoning capabilities in complex transport systems. In addition, \textcite{brasoveanuExplainableMobilityPrediction2025} present an explainable mobility prediction framework in which outputs from graph neural networks and statistical models are transformed into an \gls{rdf}[-based] knowledge graph. The approach employs a semantic conversion pipeline that utilises established ontologies such as the \gls{prov-o} for provenance tracking and the \gls{w3c} TIME ontology for temporal representation, alongside custom vocabularies for mobility-specific concepts. The resulting knowledge graph enables \gls{sparql}[-based] querying to analyse prediction confidence, user behaviour patterns, and model explanations. Unlike traditional ontology-driven approaches, the knowledge graph is not used to model domain knowledge directly, but rather to represent and analyse \gls{ai}[-generated] predictions in a semantically structured and explainable manner.

\paragraph{Studies Using Other Forms of Knowledge Graph}
In contrast, four of the reviewed studies employ knowledge graphs in a more data-driven or structural manner, without adopting formal ontology languages or semantic web standards. For example, \textcite{huIndividualTravelKnowledge2022a} constructs an individual travel knowledge graph using multimodal data sources such as smart card transactions, \gls{gps} data, and transport networks. The resulting graph captures spatio-temporal travel patterns and is combined with machine learning techniques to identify commuter behaviour. The approach here is primarily data-driven and does not incorporate formal semantic modelling. As a result while it is effective for analysing travel patterns and supporting machine learning tasks, it does not provide the semantic interoperability reasoning capabilities, or shared conceptual framework that is associated with ontology driven knowledge graphs. A different application is presented by \textcite{youFederatedPlatformEnabling2023a}, which introduces a federated platform that connects devices, data, and functions through a knowledge graph to support smart mobility services such as trip planning. Here, the knowledge graph is manually constructed to orchestrate system components by representing them as nodes and defining their interactions. While effective for system integration, the approach does not employ \gls{w3c} standards. Consequently, interoperability and semantic consistency beyond the specific platform may be more difficult to achieve.

In the context of sustainability and environmental analysis, \textcite{yanResearchSustainableDevelopment2025a} proposes a framework for evaluating urban sustainability aligned with the \glspl{sdg}. The study constructs a knowledge graph linking \gls{sdg} goals, indicators, and standards using Neo4j and Python. While the framework captures structured relationships and supports decision-making, it follows a data integration and mapping approach, possibly hindering interoperability with existing \gls{w3c}[-compliant] artefacts. 

\textcite{ahmedKnowledgeGraphBased2022a} applies graph-based representations to model trajectories for outlier (unusual travel patterns) detection in smart cities. While the study refers to its representation as a knowledge graph, the approach primarily transforms trajectory data into graph structures for similarity analysis and machine learning. It does not model domain knowledge in terms of explicit concepts or semantic relationships, and therefore aligns more closely with graph based data representations rather than knowledge graph or ontology-driven approaches.

\paragraph{Review Papers}
\textcite{rajanKnowledgeGraphsApplications2024a} provide a broad review of knowledge graph applications across smart city domains, including transport, sustainability, and urban planning. The study highlights the broader role of knowledge graphs in integrating heterogeneous urban datasets and supporting decision making across smart city domains.

\paragraph{Beyond Scope}
\textcite{tomassiHealthcareSupplyChain2025a} appears to explore healthcare supply chain management through literature-based analytical approaches, and is therefore deemed beyond the scope of this study.

\subsection{\region{DS}: Decarbonisation and Semantic Technologies} \label{sec:results-ds}
The intersection of semantic technologies and decarbonisation is sparse in our corpus, with just three associated papers (see Table~\ref{tab:results-ds}). While many papers mention ``low‑carbon'' as motivation, few combine Semantic Web technologies with a concrete decarbonisation objective. This suggests that, while semantic approaches to carbon-related modelling do exist, their application to transport decarbonisation in a standards-based Semantic Web setting remains underdeveloped.

\begin{table}[ht]
    \rowcolors{2}{gray!15}{white}
    \centering
    \caption{The studies within \region{DS} cross-checked against \Cref{tab:definitions}. These papers were reviewed by Wenshuo Tang.}
    \label{tab:results-ds}
    \resizebox{\columnwidth}{!}{%
    \begin{tabular}{lccccc}
        \toprule
        Study & \rotatebox[origin=l]{90}{Ontology} & \rotatebox[origin=l]{90}{\glsentryshort{kg}} & \rotatebox[origin=l]{90}{\glsentryshort{owl} Ontology} & \rotatebox[origin=l]{90}{\glsentryshort{rdf} \glsentryshort{kg}} & \rotatebox[origin=l]{90}{Artefact} \\
        
        \midrule
        \multicolumn{6}{c}{\textit{Using \glsxtrshort{w3c} standards}}\\
        \midrule

        \textcite{yadavSemanticBasedCarbon2023a} & \checkmark & \checkmark & \checkmark & \checkmark &  \\

        \midrule
        \multicolumn{6}{c}{\textit{Beyond scope}}\\
        \midrule

        \textcite{juEmbodiedEnergyEmergy2011} &  &  &  &  &  \\
        \textcite{wozniakEverydayProblemsVs2012a} &  &  &  &  &  \\

        \bottomrule
        
        \end{tabular}
    }
\end{table}

\paragraph{Studies Using \glsfmtshort{w3c} Standards}
\textcite{yadavSemanticBasedCarbon2023a} develop an ontology for carbon footprint analysis of food supply chains, covering animal feed, land use, and transport. The ontology comprises \num{74} classes, \num{10} object properties, \num{12} data properties, and \num{76} individuals, populated with real‑world data and evaluated via \gls{sparql} queries (e.g. finding high‑protein, low‑emission foods). Although the domain is food rather than transport, the ontology explicitly models transport‑related emissions (\texttt{foodEmissionsTransport}), and the pattern of linking activity data to emission factors is directly transferable to transport decarbonisation. A notable limitation is that the ontology remains narrowly focused on food products; the framework does not yet incorporate dynamic transport‑specific variables such as real‑time traffic conditions or vehicle‑specific emission factors, which would be necessary for direct application to transport decarbonisation scenarios. Nevertheless, this work represents one of the most complete examples identified in this review that integrates \gls{w3c} standards within a decarbonisation context.

\paragraph{Beyond Scope}
\textcite{juEmbodiedEnergyEmergy2011} discuss emissions, low-carbon economy, and biodiesel production, while referring to the concept of an ``energy ontology''. However, the ontology is used in a philosophical and ecological sense derived from Odum’s energy framework rather than as a formal \gls{owl} ontology. The work is more appropriately characterised as an energy accounting and carbon assessment study rather than a Semantic Web approach, and is thus beyond scope of our study.

Similarly, \textcite{wozniakEverydayProblemsVs2012a} present Vertoid, an application designed to help reduce domestic carbon footprints through vehicle mileage tracking, emission estimation, and eco-challenges. Although the study focuses on decarbonisation, it does not use Semantic Web technologies and focuses on user behaviour and mobile application support.

\subsection{\region{TDS}: Transport, Decarbonisation, and Semantic Technologies} \label{sec:results-tds}
Within \Cref{fig:venn3}, \region{TDS} defines the central focus of this survey article. 
The seven studies in this region provide the strongest indication of how semantic technologies may contribute to transport decarbonisation, but they also reveal how limited and uneven this evidence base currently is. Some studies employ ontologies or related semantic structures in ways that are relevant to emissions, energy systems, logistics, or electric mobility, yet the degree of direct alignment with transport, decarbonisation, and Semantic Web practice varies considerably across the Region. Taken together, these results suggest that the full integration of transport, decarbonisation, and standards-based semantic technologies remains emergent rather than mature, with relatively few studies demonstrating direct, operational support for low-carbon transport reasoning. \Cref{tab:results-tds} indicates which of studies aligns with the definitions presented in \Cref{tab:definitions}.
\begin{table}[ht]
    \rowcolors{2}{gray!15}{white}
    \centering
    \caption{The studies within \region{TDS} cross-checked against \Cref{tab:definitions}. These papers were reviewed by Daniel Mitchell.}
    \label{tab:results-tds}
    \resizebox{\columnwidth}{!}{%
    \begin{tabular}{lccccc}
        \toprule
        Study & \rotatebox[origin=l]{90}{Ontology} & \rotatebox[origin=l]{90}{\glsentryshort{kg}} & \rotatebox[origin=l]{90}{\glsentryshort{owl} Ontology} & \rotatebox[origin=l]{90}{\glsentryshort{rdf} \glsentryshort{kg}} & \rotatebox[origin=l]{90}{Artefact} \\
        
        \midrule
        \multicolumn{6}{c}{\textit{Using \glsxtrshort{w3c} standards}}\\
        \midrule

        \textcite{zhuOntologyServiceParadigm2013a} & \checkmark &  & \checkmark &  &  \\
        \textcite{mateiMultimodalTransportationOverview2021a} & \checkmark &  & \checkmark &  &  \\
        \textcite{ruizBatteryElectricVehicle2024a} & \checkmark &  & \checkmark &  & \\

        \midrule
        \multicolumn{6}{c}{\textit{Using alternative semantic technology approaches}}\\
        \midrule

        \textcite{vandersandenOntologyConsumerAcceptance2010} & \checkmark &  &  &  &  \\
        \textcite{zhaoQuantificationCarbonEmission2023a} &  & \checkmark &  &  &  \\
        \textcite{trappeyOntologybasedPatentAnalysis2024a} & \checkmark &  &  &  &  \\

        \midrule
        \multicolumn{6}{c}{\textit{Beyond scope}}\\
        \midrule
        \textcite{watanabeEstimatingQOLCar2023a} &  &  &  &  &  \\

        \bottomrule
        
        \end{tabular}
    }
\end{table}

\paragraph{Studies Using \glsfmtshort{w3c} Standards}

Carbon footprint management and labelling are addressed through an \gls{owl}[-based] ontology and service-oriented paradigm~\cite{zhuOntologyServiceParadigm2013a}. The ontology is modelled at the service and process levels, enabling the tracking of carbon emissions associated with individual processes and supporting the identification and mitigation of carbon-intensive activities. Although the approach is primarily targeted at manufacturing systems, transport is incorporated implicitly within the product lifecycle model rather than as a dedicated domain-specific component. Emissions from transport are represented at an aggregate level, corresponding to the movement of products between lifecycle stages, without explicit modelling of routing, vehicles, or traffic systems. Nevertheless, the ontology remains relevant for capturing supply chain-related emissions within carbon footprint assessments and meets the criteria of the keywords search.

\textcite{mateiMultimodalTransportationOverview2021a} present an ontology-driven framework to support the comprehensive understanding and optimisation of complex transport systems across organisational boundaries. The approach is motivated by the need to model multimodal logistics environments in an integrated manner for improved analysis and decision-making. The ontology is designed to capture the structure and interactions of transport system components, providing a holistic representation of operations spanning freight transport, warehousing, distribution, customs, and supply chain management. A real-world case study involving Ekol Logistics is used to demonstrate applicability within a large-scale, international logistics context. By formalising system knowledge, the ontology facilitates the identification of inefficiencies and supports the development of optimisation algorithms aimed at improving operational performance. This includes reducing manual intervention, enhancing cost efficiency, and minimising environmental impact. In particular, the work highlights the role of semantic modelling in enabling more effective transport optimisation strategies, with an emphasis on sustainability and emission reduction. While the ontology states optimisation of greenhouse gas emissions it is unclear whether this has been achieved or implemented further by Ekol as only partial diagrams of the ontology are provided. 

An ontology for modelling battery electric vehicle charging infrastructure to support cross-domain analysis is presented \cite{ruizBatteryElectricVehicle2024a}. The ontology is developed in accordance with \gls{fair} principles, through the integration and extension of existing ontologies. The model captures key aspects of charging infrastructure, including charging stations, vehicles, energy flows, and associated stakeholders, enabling a unified representation of the domain. By leveraging semantic web technologies, the ontology aims to facilitate interdisciplinary research and data integration across transport, energy, and infrastructure systems. However, the proposed ontology remains at the design stage and has not been validated through real-world deployment or application, limiting its demonstrated practical impact. Additionally, the ontology does not capture decarbonisation metrics in comparison against other vehicle technologies, such as against combustion engines in petrol, diesel or potential fuels of the future such as hydrogen. 

\paragraph{Studies Using Alternative Semantic Technology Approaches}
\textcite{vandersandenOntologyConsumerAcceptance2010} present an ontology for the formal representation of consumer acceptance within socio-technical energy systems. The ontology is designed to provide a shared semantic framework that facilitates interoperability between engineering and social science domains by standardising key concepts and their relationships. Core entities including agents, technologies, and acceptance dimensions, along with associated constructs such as attitudes toward adoption, adoption intention, and adoption pathways are considered. Through this structured representation, the ontology enables systematic analysis of consumer behaviour and supports the evaluation of socio-technical interventions. The proposed framework is particularly applicable to the study of technology adoption and policy impact in the energy domain, including renewable energy technologies (e.g. solar photovoltaic systems, heat pumps) and energy efficiency measures (e.g. building insulation). It further allows for the examination of policy instruments, such as subsidies and marketing strategies, in influencing public acceptance.

\textcite{zhaoQuantificationCarbonEmission2023a} proposes a method to quantify the value of \gls{ev} technologies using techniques such as natural language processing, knowledge graphs, and a Bert-BiLSTM-Attention-CRF model, and convert that value into carbon credits to support technology transfer and emissions reduction, particularly in developing countries. It also establishes a framework for carbon credit valuation based on emissions baselines and market factors, combining these with a CRITIC weighting method (a method which avoids the influence of human subjective factors such as technical experience) to inform international carbon trading strategies. 

\textcite{trappeyOntologybasedPatentAnalysis2024a} presents a survey of bike-sharing patents; however, despite referencing ontology-based approaches, the analysis is primarily based on keyword-driven patent mining, not formal semantic modelling.

\paragraph{Beyond Scope}
\textcite{watanabeEstimatingQOLCar2023a} estimates urban quality of life using point-of-view camera data and deep neural network-based image processing, without incorporating semantic web technologies or addressing decarbonisation objectives.

\section{Discussion} \label{sec:discussion}
Following the analysis presented in \Cref{sec:results}, this section discusses the key findings from our review of the literature, summarising the emerging research trends to answer \Cref{rq4}. The discussion is structured around four main findings: \Cref{sec:discussion_decarb} discusses the limited extent to which decarbonisation is addressed in existing work; \Cref{sec:discussion_misnomers} examines terminological drift within the field; \Cref{sec:discussion_artefacts} highlights the limited availability of reusable artefacts; and \Cref{sec:discussion_digital_twins} considers the emerging role of digital twins across the three thematic pillars. 

Removing studies we deemed out-of-scope in \Cref{sec:results} leaves \num{48} studies, on which we base our statistics in this section. \Cref{tab:summary} presents a summary of the findings of these \num{48} studies.
\begin{table*}[ht!]
    \centering
    \caption{A summary of the \num{48} in-scope cross-checked studies from each region. Subthemes are also included for \region{TS}.}
    \label{tab:summary}
\begin{tabular}{llrrrrrrrrrrr}
\toprule
Region & Subtheme & $N$ & \multicolumn{2}{c}{\rotatebox[origin=l]{90}{Ontology}} & \multicolumn{2}{c}{\rotatebox[origin=l]{90}{Knowledge Graph}} & \multicolumn{2}{c}{\rotatebox[origin=l]{90}{\glsxtrshort{owl} Ontology}} & \multicolumn{2}{c}{\rotatebox[origin=l]{90}{\glsxtrshort{rdf} Knowledge Graph}} & \multicolumn{2}{c}{\rotatebox[origin=l]{90}{Artefact}} \\ \cmidrule(l){4-13}
& & & $n$ & \si{\percent} & $n$ & \si{\percent} & $n$ & \si{\percent} & $n$ & \si{\percent} & $n$ & \si{\percent} \\
\midrule
\textbf{TS} &  & 41 & 35 & 85.4 & 31 & 75.6 & 26 & 63.4 & 22 & 53.7 & 8 & 19.5 \\ \addlinespace
& Semantic Web & 12 & 10 & 83.3 & 11 & 91.7 & 10 & 83.3 & 11 & 91.7 & 6 & 50.0 \\
& Ontology & 24 & 24 & 100.0 & 15 & 62.5 & 16 & 66.7 & 11 & 45.8 & 3 & 12.5 \\
& Knowledge Graph & 9 & 5 & 55.6 & 9 & 100.0 & 3 & 33.3 & 3 & 33.3 & 0 & 0.0 \\ \addlinespace
\textbf{DS} &  & 1 & 1 & 100.0 & 1 & 100.0 & 1 & 100.0 & 1 & 100.0 & 0 & 0.0 \\ \addlinespace
\textbf{TDS} &  & 6 & 5 & 83.3 & 1 & 16.7 & 3 & 50.0 & 0 & 0.0 & 0 & 0.0 \\ \addlinespace
\bottomrule
\end{tabular}
\end{table*}

\subsection{Decarbonisation is Not Widely Addressed} \label{sec:discussion_decarb}

A significant finding of this review is that decarbonisation remains weakly represented within the Semantic Web literature on transport. Of the \num{60} retained studies, \num{50} lay at the transport and Semantic Web intersection (\region{TS}), but only \num{3} addressed decarbonisation and Semantic Web together (\region{DS}), and only \num{7} spanned all three thematic pillars (\region{TDS}). Furthermore, of the \num{10} studies related to decarbonisation, only \num{7} were deemed within scope. This imbalance suggests that the literature is much more mature in semantic interoperability and transport knowledge representation than in the explicit semantic modelling of emissions reduction or low-carbon decision-making.

Across the reviewed studies, decarbonisation is often treated as a background motivation rather than as a central semantic modelling target. Many studies focus on infrastructure, logistics, public transport data, or interoperability across heterogeneous systems, but do not formally represent emissions, carbon accounting, mitigation pathways, or carbon-sensitive trade-offs in ways that can be queried or reasoned over. Even among the small number of studies retrieved at the full three-way intersection, the degree of alignment is uneven, with several papers only indirectly addressing transport, decarbonisation, or formal Semantic Web practice.

This suggests that the field currently offers more in the way of enabling semantic infrastructure than direct decarbonisation applications. Ontologies, \gls{rdf}, and \gls{lod} approaches are present, but they have rarely been applied to explicit transport decarbonisation tasks such as emissions-aware routing, multimodal transition modelling, or the integration of transport activity data with emissions factors and policy constraints. Future research should therefore move beyond invoking sustainability in general terms, and instead develop reusable Semantic Web artefacts in which decarbonisation is modelled as an explicit and operational concern.

\subsection{There is Terminological Drift Around Semantic Technologies} \label{sec:discussion_misnomers}
\begin{figure}[ht]
    \centering
    \includegraphics[width=\linewidth]{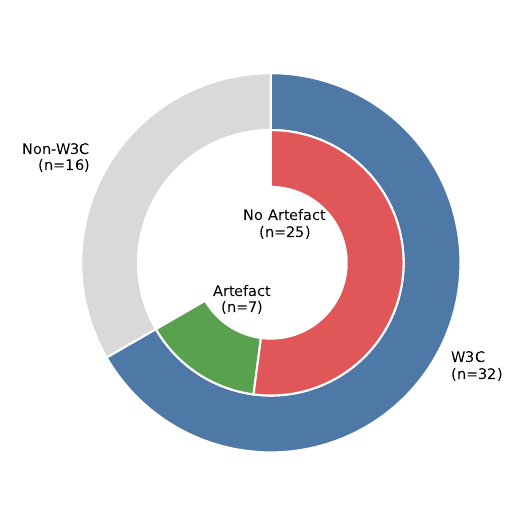}
    \caption{Only \num{32} studies of the \num{48} relevant studies used \glsxtrshort{w3c} approaches. This is defined as a study using at least one of \gls{owl} ontology or \gls{rdf} knowledge graph. Furthermore, only \num{7} of these studies present artefacts.}
    \label{fig:w3c_summary}
\end{figure}

A recurring issue in the reviewed literature is the inconsistent use of terminology around semantic technologies, ontologies, and knowledge graphs. In \Cref{tab:definitions}, we distinguish between the broader term of semantic technologies and the narrower category of Semantic Web technologies, as to separate the concerns around the approaches: the wider range of methods to represent and exploit meaning in data (semantic technologies); and the \gls{w3c} stack that enables standardised machine-readable data (Semantic Web technologies). Given this, \Cref{fig:w3c_summary} shows that \num{32} of the \num{48} in-scope studies adhere to \gls{w3c} standards.
Of these, \num{30} use \gls{owl} ontologies, and \num{23} use \gls{rdf} knowledge graphs. 

However, there are cases where this distinction is not made in the original study. This distinction matters because the two terms imply different levels of technical commitment. For example, a study may use semantic technologies without using Semantic Web technologies: \textcite{vandersandenOntologyConsumerAcceptance2010} use ontology as a shared conceptualisation for consumer acceptance in socio-technical energy systems, \textcite{curranFoundationsTransdisciplinaryEngineering2024a} introduces a transdisciplinary knowledge ontology within a wider theoretical framework, and \textcite{trappeyOntologybasedPatentAnalysis2024a} use an ontology-derived hierarchical schema to support patent mining for bike-sharing services. These are all legitimate semantic approaches, but they are not presented as \gls{owl} ontologies with queryable data or reusable web identifiers. Conversely, some studies appear to employ technologies aligned with \gls{w3c} standards, yet do not explicitly identify them as such. As a result, the boundary between semantic technologies and Semantic Web technologies is not always clearly articulated in the literature, requiring interpretation of the underlying technical artefact or figures rather than reliance on the terminology used by the authors.

A further distinction concerns the boundary between ontologies and taxonomies. Several studies present hierarchical classifications of concepts, technologies, or system components that resemble taxonomies. However, in many cases, the studies do not provide sufficient evidence to determine whether the reported model is taxonomical or ontological. Importantly, this distinction cannot be inferred solely from the technologies used to represent the model: taxonomies may be implemented using \gls{w3c} standards such as \gls{rdf}, \gls{rdfs}, \gls{skos}, or \gls{owl}, without necessarily exhibiting the semantic expressiveness typically associated with ontologies. Consequently, we adopted a conservative approach and did not classify studies into a separate taxonomy category. Nevertheless, greater clarity regarding whether a contribution is intended as a taxonomy, ontology, or another form of knowledge organisation system would improve the interpretability and comparability of studies in this domain.

In the context of transport decarbonisation, these distinctions are especially important. The motivation for using Semantic Web technologies is not merely that they provide a convenient vocabulary for describing the decarbonisation domain; rather, their value lies in supporting integration across heterogeneous decarbonisation-related datasets and models. We therefore argue that future studies in this area should be explicit about whether they are using semantic technologies in the broad sense, or Semantic Web technologies in the \gls{w3c} standards-based sense. Studies adopting \gls{w3c} standards should report which parts of the stack are used, and should be explicit about the purpose.

\subsection{The Lack of Artefacts Harms Reusability} \label{sec:discussion_artefacts}
In addition to the terminological drift addressed in \Cref{sec:discussion_misnomers}, the lack of supplementary artefacts is another issue identified in the literature. To move towards \gls{fair} principles, Semantic Web work presented as part of a study should be accompanied by publicly-accessible artefacts: this may include \gls{owl} files serialised to an \gls{rdf} format, or \gls{rdf} knowledge graph instances/endpoints. Despite this, many studies either omit these artefacts entirely or provide them in incomplete or unusable forms: \Cref{fig:w3c_summary} shows that just \num{7} of the \num{32} studies that present \gls{w3c} standard-compliant work provide a publicly available artefact.

Future research should not focus solely on the publication and description of an \gls{rdf} knowledge graph or \gls{owl} ontology within a paper, but should also provide the corresponding artefact as a distinct, accessible repository. Platforms such as Zenodo~\cite{cerndatacentreZenodo} can be used for archival, while tools like \gls{widoco}~\cite{garijoWIDOCOWizardDocumenting2017} or \gls{lode}~\cite{peroniMakingOntologyDocumentation2012} can support documentation, and \gls{oops}~\cite{poveda-villalonValidatingOntologiesOOPS2012} can assist with evaluation. Where \gls{rdf} knowledge graphs are produced, repositories should also consider including \gls{rml} or \gls{shacl} files. Such practices would align with the principles illustrated in \Cref{fig:example-workflow} and enhance the reusability of Semantic Web artefacts.

The availability of reusable artefacts is particularly important in the context of transport decarbonisation. As discussed throughout this paper, transport is a complex system-of-systems. Achieving decarbonisation therefore requires the integration and communication of data across traditionally siloed stakeholders. Semantic Web technologies provide the standards and formal semantics needed to support this interoperability, but their benefits cannot be fully realised if the underlying artefacts remain unavailable. Publicly accessible \gls{owl} ontologies, \gls{rdf} knowledge graphs, \gls{rml} mappings, and \gls{shacl} validation rules enable researchers and practitioners to reuse, extend, and link existing work, rather than repeatedly developing isolated solutions. In turn, this supports the gradual development of a broader \gls{lod} ecosystem for transport, providing a shared semantic foundation upon which data-driven decarbonisation strategies can be built. Without such openness and reuse, the risk is that semantic models remain fragmented, limiting their ability to support the coordinated actions required to decarbonise transport at scale.

\subsection{Digital Twins and Semantic Web Technologies is a Promising
Field} \label{sec:discussion_digital_twins}
An emerging trend in the reviewed literature is the convergence between \textit{digital twins} and Semantic Web technologies in transport. Digital twins~\cite{almasanNetworkDigitalTwin2022} and intelligent transport systems are becoming important research directions, but their effective deployment -- as with transport -- depends on the ability to integrate heterogeneous data, models, and domain concepts across organisational boundaries. These requirements align closely with the strengths of \gls{w3c}[-compliant] approaches.

The reviewed studies suggest that this convergence is beginning to materialise in practice. \textcite{martinezVRUSafetyRoad2025a} for example, present an ontology-driven digital twin for road intersections, using \gls{owl}[-based] modelling to integrate static infrastructure and dynamic traffic elements for visualisation, simulation, and risk analysis. More broadly, \textcite{xiaStudyCityDigital2022b} identify ontology-based integration between \gls{bim} and \gls{gis} as a key component of city digital twin research for sustainable smart cities. Together, these studies indicate that Semantic Web technologies are increasingly being positioned as enabling components of digital twin architectures rather than as purely stand-alone knowledge-engineering artefacts.

\section{Future Research Trends} \label{sec:discussion-future-research}
Based on themes identified in the survey, we now describe future research challenges in achieving semantic-enabled transport decarbonisation. This directly addresses \Cref{rq5}.

\subsection{Automated and Assisted Ontology Engineering}
While Semantic Web technologies have significant promise for transport decarbonisation, a major challenge is the expertise required to create and maintain them~\cite{lambrixCompletingDebuggingOntologies2023}. A key future research challenge is the need to make the workflow described in Figure~\ref{fig:example-workflow} more accessible.

Recent progress in \glspl{llm} may provide useful support for ontology engineering, particularly in tasks concerned with capturing, structuring, and verbalising domain knowledge. Traditional ontology development typically involves sustained interaction between ontology engineers and domain experts, and this process can be time-consuming. \Glspl{llm} do not remove the need for expert validation or formal modelling, but they may assist with specific parts of the workflow, such as drafting competency questions~\cite{alharbiInvestigatingOpenSource2024}, ontology alignment~\cite{lushneiLargeLanguageModels2026}, supporting ontology learning~\cite{magniniActivelyLearningTerminologies2025,zhaoSubsumptionSatisfiabilityLLMassisted2026}, reasoning tasks~\cite{yangLargeLanguageModel2026}, and improving ontology-based retrieval through \gls{rag}[-based] methods~\cite{zhouWhatBreaksKnowledge2026}. A promising direction for future work is therefore to examine how \gls{llm}[-assisted] methods can be combined with ontology engineering practices in a controlled way, especially for cross-area infrastructure systems where interoperability requirements are complex and domain knowledge is distributed across several sectors.

\subsection{Inference and Cause-Effect Analysis}
Beyond interoperability, a common semantic framework for transport decarbonisation also presents the opportunity for more detailed and far reaching \textit{inference}. Recent work has shown that a semantic approach to root cause and cause-effect analysis supports automation of issue identification and mitigation~\cite{chraimGraphicalCausalReasoning2026}.

In the context of transport decarbonisation, such an approach can support data-based exploration and analysis of key emission challenges, as well as future priority target areas of intervention backed by formally described and document relationships.

\subsection{Expanded Coupled Domains}
Just as semantic technologies can address the complex coupled nature of the transportation sector, they can also support expansion to other domains, in particular the energy sector. Transport decarbonisation is increasingly understood as a coupled energy--transport problem: electric vehicle charging, vehicle-to-grid services, smart mobility, and demand-side response all depend on the coordinated operation of transport and energy systems~\cite{songShapingFutureLowcarbon2022}. However, like transport, data and models in these domains remain siloed across grid operators, distribution system operators, charge-point operators, transport authorities, and mobility platforms, hindering system-level optimisation and policy analysis.

\section{Conclusion} \label{sec:conclusion}
To understand the role of semantic technologies in supporting transport decarbonisation, this paper conducted a scoping review at the intersection of semantic technologies, transport, and decarbonisation. Our results show that of the \num{222} studies identified through a Scopus query, only \num{60} were relevant to the scope of this review, with just \num{7} situated at the intersection of all three thematic pillars.

Most reviewed studies use Semantic Web technologies to integrate heterogeneous transport data. \gls{owl}, \gls{rdf}, \gls{sparql}, and related \gls{w3c} standards have been applied to describe transport networks, services, logistics systems, urban mobility data, charging infrastructure, safety scenarios, and smart city applications. In some cases, these technologies are also used to support reasoning over transport knowledge. However, the primary focus of most studies is not carbon-aware operational decision-making. Decarbonisation is frequently cited as part of the broader motivation for the research but is less commonly represented directly within the semantic models themselves, thus leaving a major research gap for future work.

The review also highlights inconsistencies in terminology across the literature, including differing uses of terms such as \textit{ontology} and \textit{knowledge graph}. This lack of consistency may limit interoperability, reuse, and the broader application of semantic technologies in support of transport decarbonisation. Furthermore, relatively few studies make their semantic artefacts publicly available. We highlight that future work must explicitly state intentions, utilise existing standards, and publish their work to adhere to the \gls{fair} principles.

We also observe a growing trend towards the use of digital twins within this research area. To support future work on semantically enabled transport decarbonisation digital twins, we identify several promising directions for further research, including automated ontology engineering, inference and cause-effect analysis, and coupling with domains such as energy to provide a connected web of data.

Overall, Semantic Web technologies provide a promising foundation for transport decarbonisation research, but their potential has yet to be fully realised. The next step is to move beyond isolated applications towards reusable, \gls{w3c}-standards-based semantic resources that support transparent data integration, analysis, and decision-making in real-world transport decarbonisation contexts.

\section*{\Glsfmtshort{credit} Author Contribution Statement}

\textbf{Joseph~D.~Preece} 
Conceptualization, 
Methodology, 
Software, 
Validation, 
Investigation, 
Data Curation, 
Writing - Original Draft, 
Writing - Review \& Editing, 
Visualization,
Supervision,
Project administration.
\textbf{Navya~Nayak} 
Conceptualization,
Methodology,
Validation, 
Investigation,
Writing - Original Draft.
\textbf{Haoruo~Zhao} 
Conceptualization,
Methodology, 
Validation, 
Investigation,
Writing - Original Draft.
\textbf{Daniel~Mitchell} 
Conceptualization,
Methodology, 
Validation, 
Investigation,
Writing - Original Draft.
\textbf{Wenshuo~Tang} 
Conceptualization,
Investigation,
Writing - Original Draft.
\textbf{Duncan~Guthrie} 
Investigation,
Writing - Original Draft
\textbf{Heather~Steele} 
Writing - Review \& Editing.
\textbf{David~Flynn} 
Writing - Review \& Editing, 
Supervision,
Project administration,
Funding acquisition.
\textbf{Philip~Greening} 
Writing - Review \& Editing, 
Project administration,
Funding acquisition.
\textbf{Bernadin~Namoano} 
Conceptualization.
\textbf{Paul~Harvey} 
Conceptualization, 
Methodology, 
Writing - Review \& Editing, 
Supervision, 
Project administration,
Funding acquisition.

\section*{Conflict of Interest Statement}
One of the papers included in the survey corpus was authored by Joseph~D.~Preece~\cite{preeceLeveragingOntochainsDistributed2024b}, who is the lead author of the present study. To mitigate any potential conflict of interest and reduce bias in the review process, this paper was independently assigned to a different author, Haoruo Zhao, for analysis. Joseph~D.~Preece was not responsible for reviewing or assessing his own work as part of the survey.

\section*{Funding Sources}
This work is supported by \pgls{epsrc} grant \href{https://gtr.ukri.org/projects?ref=EP%2FZ533221%2F1}{\textit{TransiT -- Digital Twinning Research Hub for Decarbonising Transport}} (EP/Z533221/1).

\section*{Declaration of Generative \glsfmtshort{ai} Use}
The authors used Microsoft Copilot Enterprise during the preparation of this manuscript. Generative AI was used in three specific ways: (i) to assist in the development of the transport use case presented in \Cref{sec:background}; (ii) to support the identification and grouping of subsets of Scopus-indexed keywords as part of the literature review methodology, as described in \Cref{sec:methods}; and (iii) to assist with copy-editing and language review of the manuscript. All AI-generated outputs were critically reviewed, verified, and revised by the authors prior to inclusion. The authors take full responsibility for the content of this manuscript and confirm that all research decisions, analyses, interpretations, conclusions, and intellectual contributions are their own.

\renewcommand*{\glossarypreamble}{\footnotesize}
\glssetwidest{GenCLOn} 
\printglossary[type=\acronymtype, style=alttree, title=Abbreviations]

\printbibliography


\appendix

\section{Filtered Scopus Indexed Keywords} \label{sec:filtered-keywords}

\begin{lstlisting}[style=list, caption={Scopus keywords related to Semantic Web technologies.}, label={lst:keywords-semantic-web}]
SEMANTIC_WEB_TERMS = [
    "Semantic Web",
    "Semantic Web technology",
    "Semantic technologies",
    "Semantic Web rule language (SWRL)",
    "RDF",
]
\end{lstlisting}

\begin{lstlisting}[style=list, caption={Scopus keywords related to ontology.}, label={lst:keywords-ontology}]
ONTOLOGY_TERMS = [
    "Ontology",
    "Ontology's",
    "Ontology-based",
    "Formal ontology",
    "Core ontology",
    "Smart city ontology",
    "Emergy ontology",
    "Ontology mapping",
    "Ontology design",
    "Ontology engineering",
    "Ontology-guided modeling",
    "Knowledge ontology",
]
\end{lstlisting}

\begin{lstlisting}[style=list, caption={Scopus keywords related to knowledge graphs.}, label={lst:keywords-kg}]
KNOWLEDGE_GRAPH_TERMS = [
    "Knowledge graphs",
    "Knowledge graph",
]
\end{lstlisting}

\begin{lstlisting}[style=list, caption={Scopus keywords related to \glsxtrlong{lod}.}, label={lst:keywords-lod}]
LINKED_DATA_TERMS = [
    "Linked open datum",
    "Linked open data (LOD)",
    "Linked datum",
    "Linked data",
    "Linked data fragment",
]
\end{lstlisting}

\begin{lstlisting}[style=list, caption={Scopus keywords related to semantic technologies.}, label={lst:keywords-semantic}]
COMBINED_SEMANTIC_TERMS = (
    ONTOLOGY_TERMS + 
    SEMANTIC_WEB_TERMS + 
    KNOWLEDGE_GRAPH_TERMS + 
    LINKED_DATA_TERMS
)
\end{lstlisting}

\begin{lstlisting}[style=list, caption={Scopus keywords related to transport.}, label={lst:keywords-transportation}]
TRANSPORTATION_TERMS = [
    "Urban transportation",
    "Motor transportation",
    "Roads and streets",
    "Sustainable mobility",
    "Traffic congestion",
    "Vehicles",
    "transportation planning",
    "urban transport",
    "Intelligent vehicle highway systems",
    "transportation system",
    "Street traffic control",
    "Railroads",
    "Transportation",
    "Railroad transportation",
    "railway transport",
    "Intelligent transportation systems",
    "Traffic control",
    "Sustainable transportation",
    "Urban mobility",
    "Highway traffic control",
    "Transportation system",
    "Mass transportation",
    "Public transport",
    "Sustainable transport",
    "Public transportation",
    "transportation infrastructure",
    "traffic management",
    "Transport infrastructure",
    "traffic and transport",
    "Freight transportation",
    "Public transit",
    "transportation safety",
    "Autonomous vehicles",
    "mobility",
    "Autonomous Vehicles",
    "Road network",
    "maritime transportation",
    "Transportation routes",
    "Urban freight transport",
    "Traffic signals",
    "Light rail transit",
    "Smart mobility",
    "Advanced traffic management systems",
    "Smart transportation",
    "Vehicle detection",
    "Mobility datum",
    "railway",
    "Multimodal transportation",
    "Mobility",
    "Road extraction",
    "Autonomous vehicle technologies",
    "public transport",
    "Electric vehicles",
    "road transport",
    "Road transport systems",
    "Railroad accidents",
    "Air transportation",
    "Micro-mobility",
    "intelligent transportation system",
    "Urban transportation systems",
    "Transport performance",
    "Transport analysis",
    "Transportation distribution",
    "traffic emission",
    "Transport emissions",
    "Transport modelling",
    "Transportation process",
    "Transport operations",
    "Automotive transport",
    "Public transit networks",
    "Road type",
    "Vulnerable road user",
    "Road users",
    "Road intersections",
    "transportation",
    "Traffic demands",
    "Transportation strategies",
    "air traffic control",
    "Rapid transit",
    "road construction",
    "Carrier mobility",
    "Transportation mode prediction",
    "Transportation mode",
    "Vehicles detection",
    "Traffic surveillance",
    "Traffic networks",
    "transport at the cellular level",
    "Road monitoring",
    "Non-motor vehicle",
    "Motorized vehicles",
    "Motor vehicle",
    "Nonmotorized transportation",
    "Vehicle locating systems",
    "Urban rail transit",
    "Human mobility",
    "Emergency vehicles",
    "Light rail transit (LRT)",
    "Public transport management",
    "protein transport",
    "Intermodal transportation",
    "Traffic",
    "Road transportation",
    "Intermodal transport",
    "public transportation",
    "Traffic lanes",
    "Multi-modal transportation systems",
    "Transport strategies",
    "Smart transport strategy",
    "electric vehicle",
    "Transportation planning",
    "Transport resources",
    "transportation mode",
    "transportation technology",
    "cycle transport",
    "Transport cards",
    "Private transportation",
    "Road traffic",
    "Vehicle system",
    "transportation policy",
    "Traffic safety",
    "Traffic management strategies",
    "Traffic management",
    "Air traffic control",
    "freight transport",
    "Transport process",
    "Intelligent transport systems",
    "Urban transit system",
    "Public transit behavior",
    "Road traffic injuries",
    "Road fatalities",
    "Gravel road detection and extraction",
    "Gravel road condition monitoring",
    "Forest road networks",
    "Forest and rural road network monitoring",
    "transportation development",
    "Urban traffic management",
    "Transport operators",
    "Traffic signal control",
    "Mobility service",
    "Gravel roads",
    "Road detection",
    "Road condition monitoring",
    "Traffic speed",
    "Traffic prediction",
    "Transport security",
    "Traffic restrictions",
    "Public transportation services",
    "patient transport",
    "Sustainable transportation systems",
    "Electric Vehicles (EVs)",
    "Electric mobility",
    "Real world road networks",
    "Vehicle efficiency",
    "Transportation sustainability",
    "Transportation efficiency",
    "Vehicles collision",
    "Unmanned aerial vehicles (UAV)",
    "unmanned vehicle",
    "Electric vehicle charging infrastructures",
    "Battery-electric vehicles",
    "traffic congestion",
    "Road traffic management",
    "social mobility",
    "Futuristic vehicle",
    "Electric vehicle charging",
    "Road construction",
    "Railway transportation",
    "Railway track",
    "Global mobility",
    "Railroad tracks",
    "Transport networks",
    "Transport",
    "Seamless mobility",
    "Road pixels",
    "Automatic road extraction",
    "Transportation services",
    "Transportation fleet",
    "Transport and mobilities",
    "Transport research",
    "Railway construction",
    "Heavy haul railway",
    "Urban transit",
    "Mobility predictions",
    "Mobility graphs",
    "Mobility behavior",
    "Explainable mobility graph",
    "Taiwan high speed rails",
    "High speed rail",
    "Railway infrastructure",
    "Transportation infrastructures",
    "Vehicle technology",
    "Production and transportations",
    "Air transport",
    "Employee mobility",
    "Mobility budget"
]
\end{lstlisting}

\begin{lstlisting}[style=list, caption={Scopus keywords related to decarbonisation.}, label={lst:keywords-decarbonisation}]
DECARBONISATION_TERMS = [
    "Carbon dioxide emissions",
    "carbon emission",
    "Carbon footprint",
    "CO2 emissions",
    "Decarbonisation",
    "Low-carbon",
    "Low-carbon emissions",
    "Low-carbon energy systems",
    "Low carbon energies",
    "Low carbon economy",
    "carbon utilization",
    "Carbon emissions",
    "Carbon credits"
]
\end{lstlisting}

\end{document}